\documentclass[sigconf]{acmart}

\copyrightyear{2023}
\acmYear{2023}
\setcopyright{rightsretained}
\acmConference[KDD '23]{Proceedings of the 29th ACM SIGKDD Conference on Knowledge Discovery and Data Mining}{August 6--10, 2023}{Long Beach, CA, USA}
\acmBooktitle{Proceedings of the 29th ACM SIGKDD Conference on Knowledge Discovery and Data Mining (KDD '23), August 6--10, 2023, Long Beach, CA, USA}
\acmDOI{10.1145/3580305.3599288}
\acmISBN{979-8-4007-0103-0/23/08}

\makeatother
\AtBeginDocument{%
	\providecommand\BibTeX{{%
			\normalfont B\kern-0.5em{\scshape i\kern-0.25em b}\kern-0.8em\TeX}}}
\usepackage{subcaption}
\usepackage{flushend}
\usepackage{balance}
\usepackage{lineno}
\usepackage{amsmath,amsfonts}
\usepackage{algorithm}
\usepackage{algorithmic}
\usepackage{graphicx}
\usepackage{textcomp}
\usepackage{xcolor}
\usepackage{amsthm}
\usepackage{url}
\usepackage{float}
\usepackage{multirow}
\usepackage{multicol}
\usepackage{color}
\usepackage{bm}
\usepackage{bbm}
\usepackage{enumitem}
\usepackage{balance}

\newtheorem{definition}{Definition}

\usepackage{pifont}

\newcommand{\X}{\mathbf{X}}

\newcommand{\bD}{\mathbf{D}}
\newcommand{\bA}{\mathbf{A}}

\newcommand{\bs}{\mathbf{s}}

\newcommand{\bS}{\mathbf{S}}

\newcommand{\bH}{\mathbf{H}}
\newcommand{\bh}{\mathbf{h}}

\newcommand{\bX}{\mathbf{X}}

\usepackage{array}
\newcolumntype{L}[1]{>{\raggedright\let\newline\\\arraybackslash\hspace{0pt}}m{#1}}
\newcolumntype{C}[1]{>{\centering\let\newline  \\\arraybackslash\hspace{0pt}}m{#1}}
\newcolumntype{R}[1]{>{\raggedleft\let\newline \\\arraybackslash\hspace{0pt}}m{#1}}

\usepackage{ bbold }
\DeclareMathOperator*{\argmax}{argmax}
\DeclareMathOperator*{\argmin}{argmin}



\begin{document}
\title{Contrastive Meta-Learning for Few-shot Node Classification}

\author{Song Wang}
\authornote{Equal contribution}
\affiliation{%
  \institution{University of Virginia} \country{}}
\email{sw3wv@virginia.edu}
	
	\author{Zhen Tan}
\affiliation{%
  \institution{Arizona State University} \country{}}
\email{ztan36@asu.edu}
\authornotemark[1]

				\author{Huan Liu}
\affiliation{%
  \institution{Arizona State University} \country{}}
\email{huanliu@asu.edu}

					\author{Jundong Li}
\affiliation{%
  \institution{University of Virginia} \country{}}
\email{jundong@virginia.edu}

\begin{abstract}
    Few-shot node classification, which aims to predict labels for nodes on graphs with only limited labeled nodes as references, is of great significance in real-world graph mining tasks. 
    To tackle such a label shortage issue, existing works generally leverage the meta-learning framework, which utilizes a number of \textit{episodes} to extract transferable knowledge from classes with abundant labeled nodes and generalizes the knowledge to other classes with limited labeled nodes. In essence, the primary aim of few-shot node classification is to learn node embeddings that are generalizable across different classes. To accomplish this, the GNN encoder must be able to distinguish node embeddings between different classes, while also aligning embeddings for nodes in the same class. 
    Thus, in this work, we propose to consider both the intra-class and inter-class generalizability of the model. 
    We create a novel \emph{contrastive meta-learning} framework on graphs, named COSMIC, with two key designs. First,
    we propose to enhance the intra-class generalizability by involving a \textit{contrastive two-step optimization} in each episode to explicitly align node embeddings in the same classes.
    Second, we strengthen the inter-class generalizability by generating hard node classes via a novel \textit{similarity-sensitive mix-up} strategy.
    Extensive experiments on few-shot node classification datasets verify the superiority of our framework over state-of-the-art baselines. Our code is provided\footnote{\href{https://github.com/SongW-SW/COSMIC}{https://github.com/SongW-SW/COSMIC}}.
\end{abstract}

\maketitle

\section{Introduction}
The task of node classification aims at learning a model to assign labels for unlabeled nodes on graphs~\cite{kipf2017semi,huang2020graph,ding2020inductive}. In fact, many real-world applications can be formulated as the node classification task~\cite{liu2021relative,tan2022simple}. 
For example, in social media networks~\cite{ding2019interactive,qi2011exploring} such as Facebook, where each node represents a user and edges represent friendship relations, a classification model is tasked to predict the preferences and interests of users based on their profiles and user relations.
Recently, Graph Neural Networks (GNNs)~\cite{wu2020comprehensive,velivckovic2017graph} have shown remarkable advantages in learning node representations and predicting node labels based on the learned representations.
Nevertheless, GNNs generally require a considerable number of labeled nodes to ensure the quality of learned node representations~\cite{zhou2019meta}. That being said, the performance of GNNs will severely degrade when the number of labeled nodes is limited. In practice, it often remains difficult to acquire sufficient labeled nodes for each class~\cite{ding2020graph}.
For example, GNNs are widely used in the task to identify users according to various topics~\cite{wu2020comprehensive}. However, usually only a few users in social networks are known to be associated with newly formed topics.
Then, the trained GNNs can easily encounter a significant performance drop.
Hence, there is a surge of research interests aiming at performing node classification with only limited labeled nodes as references, known as \emph{few-shot node classification}.

To tackle the few-shot node classification problem, existing works have demonstrated the effectiveness of the meta-learning strategy~\cite{liu2021relative,huang2020graph,ding2020inductive}. In general, these works first extract transferable knowledge from classes with abundant labeled nodes (i.e., \emph{meta-training classes}). Then the learned knowledge is generalized to other classes with limited labeled nodes (i.e., \emph{meta-test classes}). Particularly, these works introduce the conception of \textit{episode}~\cite{finn2017model} for the training phase to episodically emulate each target meta-test task in the evaluation phase. More specifically, in each episode, a meta-task is sampled on the graph to train the GNN model: a few labeled nodes (i.e., support set) are sampled from the meta-training classes as references for classifying the test nodes (i.e., query set) sampled from the same classes. 
By training on multiple episodes, the GNN model learns a shared node embedding space across meta-training and meta-test classes~\cite{ding2020graph,huang2020graph,liu2021relative}. 
In essence, the key to improving the performance of GNNs on meta-test classes with limited labeled nodes is to learn generalizable node embeddings~\cite{ding2020graph,liu2021relative,tan2022transductive}. In this work, we unprecedentedly propose to effectively learn generalizable node embeddings by considering two facets: intra-class and inter-class generalizability. In particular, intra-class generalizability refers to the ability of models in aligning embeddings for nodes in the same class, while inter-class generalizability measures the ability in distinguishing node embeddings among different classes. Although these two properties are crucial in learning node embeddings, due to the distinct challenges posed by the complex graph structures, it is non-trivial to guarantee them.
    Concretely, there are two key challenges in learning generalizable node embeddings for few-shot node classification. Firstly, achieving intra-class generalizability is difficult on graphs. This involves learning similar embeddings for nodes within the same class, which is crucial for classifying meta-test classes that are unseen during meta-training. However, existing methods mainly focus on distinguishing node labels in each episode and do not encourage learning similar intra-class node representations. Additionally, neighboring nodes contain crucial contextual structural knowledge for learning similar intra-class node embeddings. However, the existing strategy of sampling individual nodes in each episode fails to capture such structural information, resulting in limited intra-class generalizability. Secondly, inter-class generalizability is not guaranteed on graphs. In few-shot node classification, the model must be able to classify nodes in a variety of unseen meta-test classes. However, the meta-training classes may be insufficient or too easy to classify, resulting in a lack of capability to classify various unseen classes and low inter-class generalizability.
    

    To tackle the aforementioned challenges regarding the generalizability of learned node embeddings, we propose a novel contrastive meta-learning framework COSMIC for few-shot node classification. Specifically, our framework tames the challenges with two essential designs.
    (1) To enhance the intra-class generalizability of the GNN model, we propose to incorporate graph contrastive learning in
    each episode. As a prevailing technique used in graph representation learning, graph contrastive learning has proven to be effective in achieving comprehensive node representations~\cite{you2020graph,hassani2020contrastive}. 
    Inspired by those works, to enhance the intra-class generalizability, we propose a two-step optimization in each episode. For the first step, we conduct graph contrastive learning on nodes in the support set to update the GNN model,
    and we propose to utilize subgraphs to represent nodes to incorporate structural knowledge.
    For the second step, we leverage this updated GNN to perform classification on nodes in the query set and compute the classification loss for further updating the GNN model.
    In this way, the GNN model is forced to learn similar intra-class node embeddings via the proposed contrastive meta-learning strategy and updated on the query set for further intra-class generalizability.
    (2) To improve inter-class generalizability, we propose a novel similarity-sensitive mix-up strategy to generate additional classes in each episode. In particular, inter-class generalizability, i.e., the capability in distinguishing node embeddings in different meta-test classes, is difficult to learn when the meta-training classes are insufficient or not difficult enough. Thus, we utilize the classes in each episode to generate new hard classes via mix-up, where the mixing ratio is based on the similarity between nodes in different classes.
    The generated classes are then incorporated in the graph contrastive learning step.
    In this way, the model is forced to distinguish additional difficult classes to enhance inter-class generalizability. In summary, our contributions are:

\begin{itemize}[leftmargin=0.35cm]
\item We improve the meta-learning strategy for few-shot node classification from the perspective of intra-class and inter-class generalizability of the learned node embeddings.
    
\item We develop a novel contrastive meta-learning framework
that (1)~incorporates a two-step optimization based on the proposed contrastive learning strategy to improve intra-class generalizability;
(2) leverages the proposed similarity-sensitive mix-up strategy to generate hard classes for enhancing inter-class generalizability.
\item We conduct extensive experiments on four benchmark
node classification datasets under the few-shot scenario and validate the superiority of our proposed framework.
    \end{itemize}

\section{Preliminaries}
	\subsection{Problem Statement}
 In this section, we provide the formal definition for the problem of few-shot node classification. We first denote an input attributed graph as $G=(\mathcal{V},\mathcal{E},\X)$, where $\mathcal{V}$ is the set of nodes, and $\mathcal{E}$ is the set of edges. $\X\in\mathbb{R}^{|\mathcal{V}|\times d}$ denotes the node feature matrix, where $d$ is the feature dimension. Furthermore, the entire set of node classes is denoted as $\mathcal{C}$, which can be further divided into two disjoint sets: $\mathcal{C}_{tr}$ and $\mathcal{C}_{te}$, i.e., the sets of meta-training classes and meta-test classes, respectively. Specifically, $\mathcal{C}=\mathcal{C}_{tr}\cup\mathcal{C}_{te}$ and $\mathcal{C}_{tr}\cap\mathcal{C}_{te}=\emptyset$. It is noteworthy that the number of labeled nodes in $\mathcal{C}_{tr}$ is sufficient for meta-training, while it is generally small in $\mathcal{C}_{te}$~\cite{zhou2019meta,huang2020graph,ding2020graph,liu2021relative}. In this way, the studied problem of few-shot node classification is formulated as follows:
	
	\begin{definition}
	\textbf{Few-shot Node Classification:} Given an attributed graph $G=(\mathcal{V},\mathcal{E},\X)$ and a meta-task $\mathcal{T}=\{\mathcal{S}, \mathcal{Q}\}$ sampled from $\mathcal{C}_{te}$, our goal is to develop a learning model such that after meta-training on labeled nodes in $\mathcal{C}_{tr}$, the model can accurately predict labels for the nodes in the query set $\mathcal{Q}$, where the only available reference is the limited labeled nodes in the support set $\mathcal{S}$.
	\end{definition}
	
	Moreover, under the $N$-way $K$-shot setting, the support set $\mathcal{S}$ consists of exactly $K$ labeled nodes for each of the $N$ classes from $\mathcal{C}_{te}$, and the query set $\mathcal{Q}$ is also sampled from the same $N$ classes. In this scenario, the problem is called an $N$-way $K$-shot node classification problem. Essentially, the objective of few-shot node classification is to learn a model that can well generalize to meta-test classes in $\mathcal{C}_{te}$ with only limited labeled nodes as the reference. 
	
\subsection{Episodic Training}
In practice, we adopt the prevalent episodic training framework for the meta-training process, which has proven to be effective in various fields, such as few-shot image classification and few-shot knowledge completion~\cite{snell2017prototypical,finn2017model,vinyals2016matching,xiong2018one,ding2020graph}. Particularly, the meta-training process is conducted on a certain number of \textit{episodes}, each of which contains a \emph{meta-training task} emulating the structure of \emph{meta-test tasks}. The only difference is that the meta-training tasks are sampled from $\mathcal{C}_{tr}$, while the meta-test tasks are sampled from $\mathcal{C}_{te}$. In this regard, the model can keep the consistency between meta-training and meta-test. 
Moreover, many works~\cite{finn2017model,xiong2018one,ding2020graph,huang2020graph} have demonstrated the benefits of such emulation-based learning strategy for better classification performance, especially under few-shot settings. Our proposed framework generally follows this design and inherits its merits in preserving classification performance with only scarce labeled nodes. 
	 \begin{figure*}[!t]
\vspace{-0.3cm}

  \scalebox{0.94}{
  \centering
  \includegraphics[width=\linewidth]{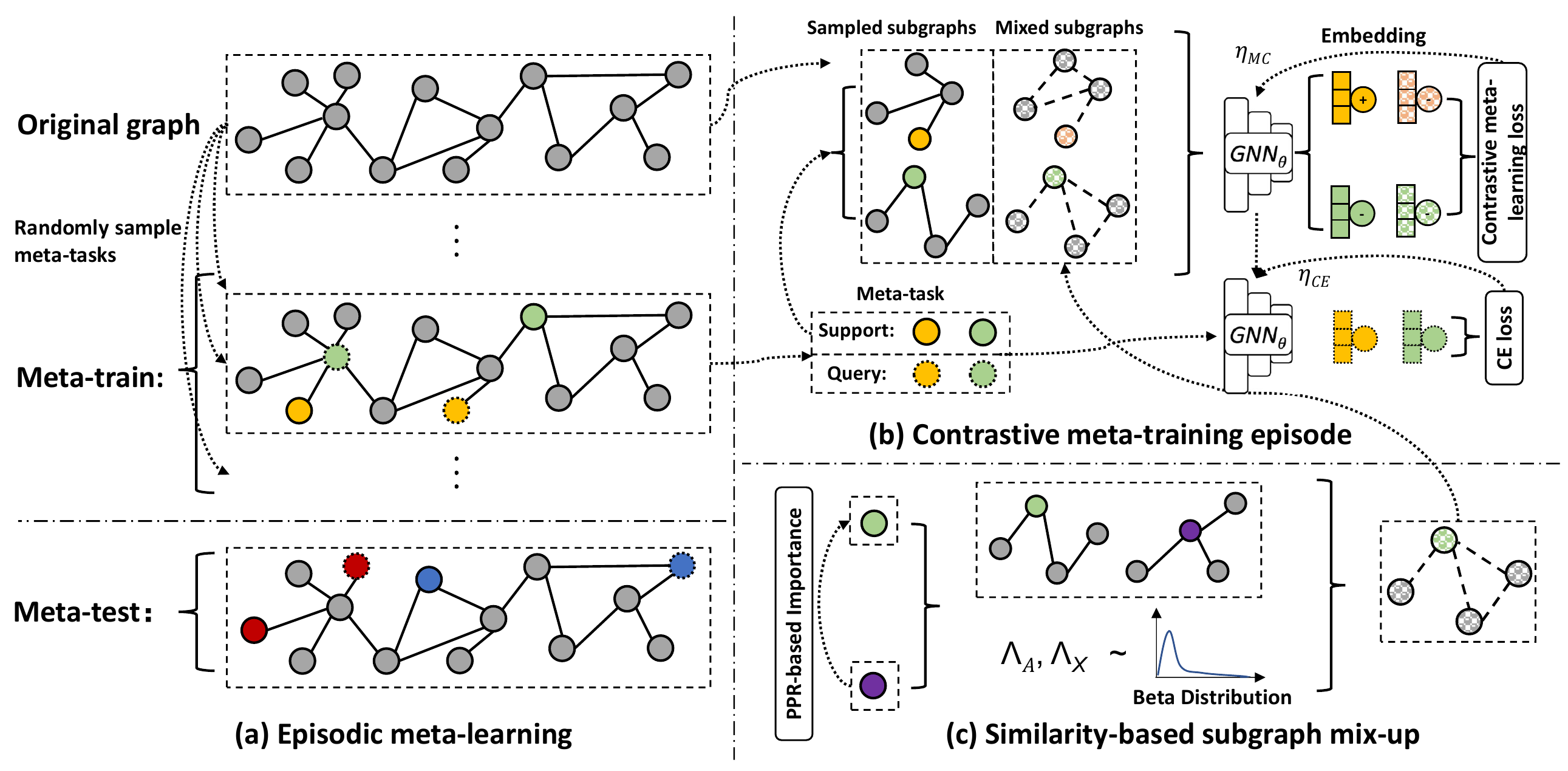}}
  \caption{The illustration of the proposed framework COSMIC under the $2$-way $1$-shot setting: (a) Episodic meta-learning framework. (b) Contrastive Meta-learning strategy during each episode. (c) similarity-sensitive subgraph mix-up strategy. Specifically, CE loss denotes the cross-entropy loss, and nodes in different colors indicate different classes.}
  \label{fig:meta-con}
\end{figure*}

\section{Proposed Framework}
	
In this section, we introduce the overall structure of our proposed framework COSMIC in detail. As illustrated in Fig.~\ref{fig:meta-con}, we formulate the problem of \emph{few-shot node classification} under the prevalent $N$-way $K$-shot episodic training framework. That being said, 
each meta-training task contains $K$ nodes for each of $N$ classes as the support set and several query nodes to be classified. Specifically, we aim to train an encoder $\text{GNN}_\theta$, parameterized by $\theta$, for classification in meta-test tasks.
To learn generalizable node embeddings, we first propose a contrastive meta-learning strategy to enhance the intra-class generalizability of $\text{GNN}_\theta$. Then we propose a similarity-sensitive mix-up strategy to introduce additional classes to improve inter-class generalizability. 
After meta-training the GNN model on a total number of $T$ meta-training tasks, we will evaluate the obtained $\text{GNN}_\theta$ on a series of meta-test tasks sampled from meta-test classes $\mathcal{C}_{te}$. Because the learned node embeddings are highly discriminative, for each meta-test task, we involve a new simple classifier (Logistic Regression in practice) for label predictions.

\subsection{Contrastive Meta-Learning on Graphs}

\subsubsection{Contrastive Meta-learning Framework}
In many of the existing few-shot node classification works~\cite{huang2020graph,liu2021relative,zhou2019meta}, the model generally encodes each node via the GNN and utilizes the learned node embeddings to distinguish classes in each meta-training task. However, such a strategy does not force the GNN to learn similar intra-class node representations, thus resulting in the lack of intra-class generalizability.
Instead, to promote the intra-class generalization to meta-test classes, we propose to integrate 
graph contrastive learning into each meta-task, so that the GNN will be trained to minimize the distance between representations of nodes in the same class in each support set. The updated GNN will output node representations for nodes in the query set for classification. In this manner, the GNN model will learn the intra-class generalizability in each meta-task.
	  	
Specifically, to construct a meta-training (or meta-test) task $\mathcal{T}$, we first randomly sample $N$ classes from $\mathcal{C}_{tr}$ (or $\mathcal{C}_{te}$). Then we randomly sample $K$ nodes from each of the $N$ classes (i.e., $N$-way $K$-shot) to establish the support set $\mathcal{S}$. Similarly, the query set $\mathcal{Q}$ consists of $Q$ different nodes (distinct from $\mathcal{S}$) from the same $N$ classes. The components of the sampled meta-task $\mathcal{T}$ can be denoted as follows: 
\begin{equation}
    \begin{aligned}
    \mathcal{S}&=\{(v_1,y_1),(v_2,y_2),\dotsc,(v_{N\times K},y_{N\times K})\},\\
    \mathcal{Q}&=\{(q_1,y'_1),(q_2,y'_2),\dotsc,(q_{Q},y'_{Q})\},\\
    \mathcal{T}&=\{\mathcal{S},\mathcal{Q}\},
    \end{aligned}
\end{equation}
where $v_i$ (or $q_i)$ is a node in $\mathcal{V}$, and $y_i$ (or $y'_i$) is the corresponding label. 
Specifically, given the meta-task $\mathcal{T}$ and its support set $\mathcal{S}$ ($|\mathcal{S}|=N\times K$) on a graph $G=(\mathcal{V},\mathcal{E},\X)$, we aim to conduct contrastive learning while incorporating supervision information to enhance intra-class generalizability. We denote the $j$-th node in the $i$-th class in $\mathcal{S}$ as $v_i^j$ and its corresponding encoded representation as $\bh_i^j$ ($i=1,2,\dotsc,N$ and $j=1,2,\dotsc,K$), which is learned by $\text{GNN}_\theta$. 
To enhance the intra-class generalizability, we propose to leverage the concept of mutual information (MI) from deep InfoMax~\cite{bachman2019learning}:
\begin{equation}
    \mathcal{L}_{i,j}=-\frac{1}{NK}\sum_{i=1}^N\sum_{j=1}^K\log\frac{\text{MI}(v_i^j, C_i)}{\sum_{k=1,k\neq i}^N \text{MI}(v_i^j, C_k)},
    \label{eq:meta-contrastive}
\end{equation}
where $C_i$ is the node set for the $i$-th class in $\mathcal{S}$, i.e., $C_i=\{v_i^j|j=1,2,\dotsc,K\}$. Moreover, $\text{MI}(v_i^j, C_i)$ denotes the mutual information between node $v_i^j$ and $C_i$. That being said, in the above equation, we aim to maximize the MI between nodes in the same class while minimizing the MI between nodes in different classes. To be specific, the MI term is implemented based on the following formula:
\begin{equation}
\begin{aligned}
    \text{MI}(v_i^j, C_k)
    =&\sum_{t=1}^V \sum_{l=1}^K \sum_{r=1}^V \exp(f_t(v_i^j)\cdot f_r(v_k^l)/\tau)\\
    &-\delta_{i,k}\sum_{t=1}^V\exp(f_t(v_i^j)\cdot f_t(v_k^j)/\tau),
\end{aligned}
\end{equation}
where $V$ is the number of views for each node, and $f_t(\cdot)$ is the $t$-th learning function, where $t=1,2,\dotsc,V$. Specifically, $f_t(\cdot)$ first transforms the input node into a view with a specific function and then learns a representation from the transformed view, where the output is a $d$-dimensional vector. A more detailed description of how to generate views by transforming nodes is given in the next subsection.
    %
	%
	  Moreover, $\mathcal{L}_{i,j}$ is the loss for $v_i^j$ in a meta-task $\mathcal{T}$, and
	  $\tau\in\mathbb{R}^+$ is a scalar temperature parameter. Here the negative samples include all other nodes in the support set, while the positive samples are other nodes that share the same class as $v_i^j$.
	  In each meta-task, the loss is computed over all nodes in the support set $\mathcal{S}$, thus forming the proposed contrastive meta-learning loss $\mathcal{L}_{MC}$,
	  \begin{equation}
	      \mathcal{L}_{MC}=\frac{1}{NK}\sum_{i=1}^N\sum_{j=1}^K\mathcal{L}_{i,j}.
	  \end{equation}
We conduct contrastive meta-learning on $T$ episodes, where each episode consists of a meta-training task $\mathcal{T}_t=\{\mathcal{S}_t,\mathcal{Q}_t\}$, $t=1,2,\dotsc,T$. With the objective $\mathcal{L}_{MC}$, we first perform one gradient descent step for $\text{GNN}_\theta$ to fast adapt it to a meta-training task $\mathcal{T}_t$:
	  	  \begin{equation}
	      \widetilde{\theta}^{(t)}\leftarrow \theta^{(t)} -\eta_{MC}\nabla_{\theta^{(t)}}\mathcal{L}_{MC}\left(\mathcal{S}_t;\theta^{(t)}\right),
       \label{eq:update1}
	  \end{equation}
   where $\mathcal{S}_t$ is the support set of the meta-training task $\mathcal{T}_t$ sampled in episode $t$, and $t\in\left\{1,2,\dotsc,T\right\}$. $\mathcal{L}_{MC}(\mathcal{S}_t;\theta^{(t)})$ denotes the contrastive meta-learning loss calculated on $\mathcal{S}_t$ with the GNN parameters $\theta^{(t)}$. $\eta_{MC}$ is the learning rate for $\mathcal{L}_{MC}$.
   For the second step of our contrastive meta-learning, we incorporate the cross-entropy loss $\mathcal{L}_{CE}$ on the query set
   via another step of gradient descent,:
	  	  \begin{equation}
	      \theta^{(t+1)}\leftarrow \widetilde{\theta}^{(t)} -\eta_{CE}\nabla_{\widetilde{\theta}^{(t)}}\mathcal{L}_{CE}\left(\mathcal{Q}_t;\widetilde{\theta}^{(t)}\right),
       \label{eq:update2}
	  \end{equation}
	  where $\mathcal{Q}_t$ denotes the query set of the meta-task $\mathcal{T}_t$ sampled in episode $t$. $\mathcal{L}_{CE}$ denotes the cross-entropy loss calculated on the query set $\mathcal{Q}_t$ with the updated GNN parameters $\widetilde{\theta}^{(t)}$. $\eta_{CE}$ is the corresponding learning rate. It is noteworthy that a fully-connected layer is used during meta-training for the cross-entropy loss. As a result, through this two-step optimization, we have conducted one episode of contrastive meta-learning to obtain the updated GNN parameters $\theta^{(t+1)}$. After training on a total number of $T$ episodes, we can obtain the final trained $\text{GNN}_\theta$ with parameters $\theta^{(T)}$.
	  It is noteworthy that different from the supervised contrastive loss~\cite{khosla2020supervised}, our contrastive meta-learning loss restricts the classification range to $N$ classes in each meta-task, and then the updated GNN will be used for classification in the query set. In such a manner, the model will be trained to fast adapt to various tasks with different classes, thus learning intra-class generalizability when the model is enforced to classify unseen nodes in the query set.
	  In addition, our design also differs from infoPatch~\cite{liu2021learning} that contrasts between support samples and query samples of different views since our contrastive meta-learning loss is specified for the support set.

\subsubsection{Subgraph Construction}

In this part, we introduce the function to generate different views (i.e., $f_t(\cdot)$) for our contrastive meta-learning framework. To incorporate more structural context in each meta-task, we propose to represent each node by the subgraph extracted from it. This is because nodes are often more correlated to their regional neighborhoods compared with other long-distance nodes~\cite{zhu2021shift,wu2022handling}. Nevertheless, directly sampling its neighborhood can potentially incorporate redundant information~\cite{huang2020graph}. Therefore, we propose to selectively sample neighborhoods as the subgraph based on the Personalized PageRank (PPR) algorithm~\cite{jeh2003scaling}. Such subgraph sampling scheme has been proven effective for many different graph learning tasks~\cite{zhang2020graph,jiao2020sub}, and we validate its effectiveness for few-shot node classification in this paper. As a result, the sampled subgraph will include the most important nodes regarding the central node, which can provide context information for each meta-task while substantially reducing the irrelevant information in neighborhoods.

Concretely, given the adjacency matrix $\bA\in\mathbb{R}^{|\mathcal{V}|\times|\mathcal{V}|}$ of graph $G=(\mathcal{V},\mathcal{E},\X)$, an importance score matrix $\bS$ is computed as follows:
\begin{equation}
    \bS=\zeta\cdot \left( \mathbf{I} - (1-\zeta)\cdot \bar{\bA}\right),
\end{equation}
where $\mathbf{I}$ is the identity matrix, and $\zeta\in[0,1]$ is an adjustable parameter. $\bar{\bA}\in\mathbb{R}^{|\mathcal{V}|\times|\mathcal{V}|}$ is the column-normalized adjacency matrix computed by $\bar{\bA}=\bA\bD^{-1}$, where $\bD\in\mathbb{R}^{|\mathcal{V}|\times|\mathcal{V}|}$ is the column-wise diagonal matrix with $\bD_{i,i}=\sum_{j=1}^{|\mathcal{V}|}\bA_{i,j}$. In this way, $\bS_{i,j}$ can represent the importance score between nodes $v_i$ and $v_j$. To select the most correlated nodes for an arbitrary node $v_i$, we extract the nodes that bear the largest important scores to $v_i$ as follows:
\begin{equation}
    \Gamma(v_i)=\{v_j|v_j\in\mathcal{V}\setminus\{v_i\},\bS_{i,j}>h_{i,K_s}\},
    \label{eq:subgraph}
\end{equation}
where $\Gamma(v_i)$ is the extracted node set for the centroid node $v_i$, and $h_{i,K_s}$ is the importance score threshold for other nodes to be selected. Specifically, to ensure a consistent size of extracted subgraphs, we define $h_{i,K_s}$ as the $K_s$-th largest entry of $\bS_{i,:}$ (with $v_i$ itself excluded), where $K_s$ is a hyperparameter. In other words, $\Gamma(v_i)$ consists of the top-$K_s$ important nodes for $v_i$ on graph $G$. In this manner, the extracted subgraph node set is denoted as $\mathcal{V}_i=\{v_i\}\cup\Gamma(v_i)$. Here the original edges in this subgraph will be kept. Thus, the edge set of the subgraph can be represented as follows:
\begin{equation}
    \mathcal{E}_i=\{ (u, v)|u\in\mathcal{V}_i, v\in\mathcal{V}_i\}.
\end{equation}
Note that this strategy can extract both the neighboring nodes of $v_i$ and other nodes that are far away and important, which could incorporate more contextual knowledge.

Based on the proposed subgraph sampling strategy, for nodes in the support set $\mathcal{S}$ ($|\mathcal{S}|=N\times K$) of meta-task $\mathcal{T}$, we can accordingly extract $N\times K$ subgraphs, denoted as $\mathcal{G}=\{G_i^j|i=1,2,\dotsc,N, j=1,2,\dotsc,K\}$. Note that $G_i^j=(\mathcal{V}_i^j, \mathcal{E}_i^j, \bX_i^j)$, where $\mathcal{V}_i^j$, $\mathcal{E}_i^j$, and $\bX_i^j$ are the node set, edge set, and feature matrix of subgraph $G_i^j$, respectively.  With the encoder $\text{GNN}_\theta$, we can obtain the representation of each node as:
\begin{equation}
    \bH_i^j=\text{GNN}_\theta\left(\mathcal{V}_i^j, \mathcal{E}_i^j, \bX_i^j\right),
\end{equation}
\begin{equation}
    f_1(v_i^j)=F_1\left(\bH_i^j\right)=F_1\left(\text{GNN}_\theta\left(\mathcal{V}_i^j, \mathcal{E}_i^j, \bX_i^j\right)\right),
\end{equation}
\begin{equation}
    f_2(v_i^j)=F_2\left(\bH_i^j\right)=F_2\left(\text{GNN}_\theta\left(\mathcal{V}_i^j, \mathcal{E}_i^j, \bX_i^j\right)\right),
\end{equation}
where $\bH_i^j\in\mathbb{R}^{|\mathcal{V}_i^j|\times d_\theta}$ denotes the learned representations of nodes in $\mathcal{V}_i^j$, and $d_\theta$ is the output dimension of the encoder $\text{GNN}_\theta$. To separately encode the local information of each node, we extract the learned representation of the central node from $\bH_i^j$ via $F_1(\bH_i^j)=\text{Centroid}\left(\bH_i^j\right)$, 
where $\text{Centroid}(\cdot)$ denotes the operation that extracts the representation of the central, i.e., the node from which the subgraph $G_i^j$ is constructed. To obtain the representation of the subgraph $G_i^j$, we adopt the meaning pooling strategy: 
$
   F_2(\bH_i^j)=\text{Mean}\left(\bH_i^j\right)
$,
where $\text{Mean}(\cdot)$ is the operation that averages all node representations in the subgraph. As a result, the context information of node $v_i^j$ is encoded into $f_2(v_i^j)$. In this manner, we can obtain the two views for node $v_i^j$, i.e., $f_1(v_i^j)$ and $f_2(v_i^j)$.


\subsection{Similarity-sensitive Mix-up}
Although we have enhanced the intra-class generalizability of GNN models via our contrastive meta-learning framework, the lack of inter-class generalizability can still lead to suboptimal performance on meta-test classes. In particular, we propose a similarity-sensitive mix-up strategy to generate additional classes in each meta-task to compensate for the potential lack of sufficient or difficult meta-training classes for inter-class generalizability. The generated classes are based on mixed subgraphs and will be incorporated into our contrastive meta-learning loss. In this way, the model is forced to distinguish between both the $N$ classes in each meta-task and the generated (unseen) classes, thus promoting inter-class generalizability. It is worth mentioning that in our framework, the extracted subgraphs all maintain the same size of $K_s$, which provides further convenience for performing subgraph-level mix-up. 

Specifically, for each of $N\times K$ nodes in the support set $\mathcal{S}$, we propose to generate a corresponding mixed subgraph.
Here we denote the set of mixed subgraphs as $\widetilde{\mathcal{G}}=\{\widetilde{G}_i^j|i=1,2,\dotsc,N,j=1,2,\dotsc,K\}$, where $\widetilde{G}_i^j$ is the mixed subgraph generated for the $j$-th node in the $i$-th class. 
For each node in $\mathcal{S}$, we first randomly sample a node $v_m$ from the input graph $G$ and generate its subgraph $G_m=(\bA_m,\bX_m)$. Then with the extracted subgraph for node $v_i^j$, i.e., $G_i^j=(\bA_i^j,\bX_i^j)$, we can perform mix-up as follows:
\begin{equation}
\label{eq:mixup}
\begin{aligned}
   \widetilde{\bA}_i^j&=\Lambda_A \circ \bA_i^j + (1-\Lambda_A)\circ \bA_m, \\
  \widetilde{\bX}_i^j&=\Lambda_X\circ  \bX_i^j + (1-\Lambda_X) \circ \bX_m,
 \end{aligned}
\end{equation}
where $\widetilde{\bA}_i^j$ and $\widetilde{\bX}_i^j$ are the mixed adjacency matrix and feature matrix for the subgraph generated from node $v_i^j$, respectively. Moreover, $\circ $ denotes the element-wise multiplication operation. $\Lambda_A\in\mathbb{R}^{K_s\times K_s}$ and $\Lambda_X\in\mathbb{R}^{K_s\times d}$ are mixing ratio matrices based on the similarity between the two subgraphs $G_i^j$ and $G_m$. 

To provide more variance for the mix-up strategy, we sample each element in $\Lambda_A$ and $\Lambda_X$, i.e., $\lambda \in [0,1]$, independently from the commonly used Beta$(\alpha_i^j,\beta)$ distribution~\cite{zhang2017mixup}:
\begin{equation}
\begin{aligned}
    \lambda_{p,q}&\sim\text{Beta}(\alpha_i^j,\beta), \forall p, q \in \{1,2,\dotsc, K_s\}, \\
    \lambda_{r,t}&\sim\text{Beta}(\alpha_i^j,\beta), \forall r\in \{1,2,\dotsc, K_s\}, \forall t\in \{1,2,\dotsc,d\}.
\end{aligned}
\end{equation}


To adaptively control the mixup procedure to generate harder instances, we further design a similarity-sensitive strategy to decide the value of the parameter $\alpha_i^j$ in the Beta distribution.
Generally, if the sampled node $v_m$ is dissimilar to node $v_i^j$, we should increase the value of $\alpha_i^j$ so that a smaller $\lambda$ is more likely to be sampled. Then, based on Eq.~\eqref{eq:mixup}, the mixed subgraph will absorb more information from a different structure, i.e., the subgraph generated from $v_m$.
Particularly, we propose to adaptively adjust the $\alpha$ value in the Beta distribution based on the Bhattacharyya distance~\cite{bhattacharyya1946measure} between node $v_i^j$ and node $v_m$ regarding their importance scores based on the Personalized PageRank algorithm. Intuitively, Bhattacharyya distance can measure the contextual relations between two embeddings in a space formed by the importance scores:
\begin{equation}
	\alpha_i^j= \sigma\left( -\ln\left(   \frac{\sum_{k=1}^{|\mathcal{V}|}\sqrt{\bs_i^j(k)\cdot\bs_m(k)}}{\sqrt{\sum_{l=1}^{|\mathcal{V}|}\bs_i^j(l)\cdot\sum_{l=1}^{|\mathcal{V}|}\bs_m(l)}}\right) \right)\cdot C,
\end{equation}
where $\bs_i^j\in\mathbb{R}^{|\mathcal{V}|}$ (or $\bs_m\in\mathbb{R}^{|\mathcal{V}|}$) denotes the importance score of node $v_i^j$ (or $v_m$), i.e., the corresponding row in $\bS$. Moreover, $\bs_i^j(k)$ denotes the $k$-th entry in $\bs_i^j$. $\sigma(\cdot)$ is the sigmoid function, and $C$ is a constant value to control the magnitude of $\alpha_i^j$.
In this way, we can obtain an adaptive value for the parameter of the Beta distribution, from which $\Lambda_A$ and $\Lambda_X$ are sampled.

Denoting the mixed subgraph as $\widetilde{G}_i^j=(\widetilde{\bA}_i^j,\widetilde{\bX}_i^j)$, we can obtain the set of mixed subgraphs $\widetilde{\mathcal{G}}=\{\widetilde{G}_i^j|i=1,2,\dotsc,N,j=1,2,\dotsc,K\}$ after performing mix-up for each node in $\mathcal{S}$. In this way, we can obtain $N$ mixed classes in addition to the original $N$ classes in each meta-task. Denoting the central node in each mixed subgraph as $\widetilde{v}_i^j$, the generated mixed classes can be represented as $\widetilde{C}_i=\{\widetilde{v}_i^j|j=1,2,\dotsc,K\}$. Then these mixed classes are used as additional classes in Eq.~(\ref{eq:meta-contrastive}), which is improved as follows:
\begin{equation}
\begin{aligned}
    \mathcal{L}_{i,j}
    &=-\frac{1}{NK}\sum_{i=1}^N\sum_{j=1}^K\log\frac{\text{MI}(v_i^j, C_i)}{\sum_{k=1,k\neq i}^N \text{MI}(v_i^j, C_k)+\sum_{k=1}^N \text{MI}(v_i^j, \widetilde{C}_k)} \\
    &-\frac{1}{NK}\sum_{i=1}^N\sum_{j=1}^K\log\frac{\text{MI}(\widetilde{v}_i^j, \widetilde{C}_i)}{\sum_{k=1,k}^N \text{MI}(\widetilde{v}_i^j, C_k)+\sum_{k=1,k\neq i}^N \text{MI}(\widetilde{v}_i^j, \widetilde{C}_k)}.
\end{aligned}
\end{equation}
In this way, the model is forced to distinguish between these $N$ additional mixed classes that are potentially more difficult than the original $N$ classes.

	\begin{algorithm}[t]
		\caption {\textsc{Learning Process of the Proposed Framework.\label{algo:process}}}
		\begin{algorithmic}[1]
			\REQUIRE A graph $G=(\mathcal{V},\mathcal{E},\X)$, a meta-test task $\mathcal{T}_{test}=\{\mathcal{S},\mathcal{Q}\}$, meta-training classes $\mathcal{C}_{tr}$, the number of meta-training episodes $T$, the number of classes in each meta-task $N$, and the number of labeled nodes for each class $K$.
			\ENSURE Predicted labels for the query nodes in $\mathcal{Q}$ of $\mathcal{T}_{test}$.
			
			// \texttt{Meta-training phase}
			\STATE $t \gets 0$ ;
			\WHILE{$t<T$}
			\STATE Sample a meta-training task $\mathcal{T}_t=\{\mathcal{S}_t,\mathcal{Q}_t\}$ from $\mathcal{C}_{b}$;
			\STATE Construct a subgraph for each node in the support set $\mathcal{S}_t$;
			\STATE Mix up constructed subgraphs based on similarities;
			\STATE Compute the representations for all subgraphs and nodes with the encoder $\text{GNN}_\theta$;
			\STATE Compute the contrastive meta-learning loss for each node based on the learned representations according to Eq. (\ref{eq:meta-contrastive});
			\STATE Update parameters of $\text{GNN}_\theta$ with the contrastive meta-learning loss on nodes in $\mathcal{S}_t$ by one gradient descent step based on Eq.~(\ref{eq:update1});

            \STATE Update parameters of $\text{GNN}_\theta$ with the cross-entropy loss on nodes in $\mathcal{Q}_t$ by one gradient descent step based on Eq.~(\ref{eq:update2});
                \STATE $t\gets t+1$;
			\ENDWHILE
			
			// \texttt{Meta-test phase}
			\STATE Construct a subgraph for each node in the support set $\mathcal{S}$ and the query set $\mathcal{Q}$;
			\STATE Compute the subgraph representations for nodes in $\mathcal{S}$ and $\mathcal{Q}$ with the trained encoder $\text{GNN}_\theta$;
			\STATE Fine-tune a simple classifier $g_\phi$ based on the subgraph representations from the support set $\mathcal{S}$ according to Eq.~(\ref{eq:fine-tune});
			\STATE Predict labels for query nodes based on the subgraph representations from the query set $\mathcal{Q}$;

		\end{algorithmic}
        \label{algorithm}
	\end{algorithm}

\subsection{Meta-test}
During meta-test, we leverage the trained encoder $\text{GNN}_\theta$ to learn a representation for each node based on its extracted subgraph. Specifically, for a given meta-task $\mathcal{T}=\{\mathcal{S},\mathcal{Q}\}$, a new simple classifier $g_\phi$ (implemented as a fully connected layer parameterized by $\phi$) is trained on the support set $\mathcal{S}$ based on the cross-entropy loss:
	\begin{equation}
	    \mathbf{p}_i=\text{Softmax}\left(g_\phi(f_q(v_i))   \right),
	\end{equation}
\begin{equation}
	    \mathcal{L}_{CE}\left(\mathcal{S};\theta^{(T)}\right)=-\frac{1}{|\mathcal{S}|}\sum\limits_{i=1}^{|\mathcal{S}|}\sum\limits_{j=1}^{N}y_{i,j}\log p_{i,j},
	    \label{eq:fine-tune_CE}
\end{equation}
where $\mathbf{p}_i\in\mathbb{R}^{N}$ is the probability that the $i$-th support node $v_i$ in $\mathcal{S}$ belongs to each class of the $N$ classes in meta-task $\mathcal{T}$. Here $|\mathcal{S}|=N\times K$ under the $N$-way $K$-shot setting. $\theta^{(T)}$ denotes parameters of $\text{GNN}_\theta$ after $T$ episodes of meta-training.
Moreover, $y_{i,j}=1$ if the $i$-th node belongs to the $j$-th class in $\mathcal{T}$, and $y_{i,j}=0$, otherwise. $p_{i,j}$ denotes the $j$-th element in $\mathbf{p}_i$. $f_q(v_i)$ is the representation of the subgraph $G_i=(\mathcal{V}_i,\mathcal{E}_i,\bX_i)$ extracted from $v_i$ based on Eq. (\ref{eq:subgraph}), which is learned by the GNN encoder:
\begin{equation}
    f_q(v_i)=F_q\left(\text{GNN}_\theta\left(\mathcal{V}_i,\mathcal{E}_i,\bX_i\right)\right).
    \label{eq:fq}
\end{equation}
It is notable that during training the classifier, the parameters of the encoder $\text{GNN}_\theta$ are fixed to ensure a faster convergence. Moreover, we additionally introduce a weight-decay regularization term $R(\phi)=\|\phi\|^2/2$. In consequence, we can achieve a classifier that is specified for the meta-task $\mathcal{T}$ based on the following objective:
\begin{equation}
    \phi^*=\argmin_\phi \mathcal{L}_{CE}(\mathcal{T};\theta,\phi)+R(\phi),
    \label{eq:fine-tune}
\end{equation}
where $\phi^*$ denotes the optimal parameters for the classifier $\text{ML}_\phi$ (In practice, we choose Logistic Regression as the classifier).
Then we conduct classification for nodes in the query set $\mathcal{Q}$ with the learned subgraph representations of these nodes. 
The label of the $i$-th query node $q_i$ is obtained by $\hat{y}_i=\argmax_j\{p^q_{i,j}\}$. Here $p^q_{i,j}$ is the $j$-the element of $\mathbf{p}^q_i$, which is learned in a similar way as Eq.~(\ref{eq:fine-tune}) and Eq.~(\ref{eq:fq}) based on the extracted subgraph of $q_i$. The process of our framework is described in Algorithm~\ref{algo:process} and illustrated in Fig.~\ref{fig:meta-con}.

    \section{Experimental Evaluations}
    To achieve an empirical evaluation of our framework COSMIC, we conduct experiments on four prevalent real-world node classification datasets with different few-shot settings, 
    including CoraFull~\cite{bojchevski2018deep}, ogbn-arxiv~\cite{hu2020open}, Coauthor-CS~\cite{shchur2018pitfalls}, and DBLP~\cite{tang2008arnetminer}. Their statistics and class splitting policy are provided in Table~\ref{tab:statistics}. More detailed descriptions are included in Appendix~\ref{app:dataset}.

	\begin{table*}[htbp]
			   					\setlength\tabcolsep{4.6pt}

		\centering
				\renewcommand{\arraystretch}{1.2}
		\caption{The overall few-shot node classification results of all methods under different settings. Accuracy ($\uparrow$) and confident interval ($\downarrow$) are in $\%$. The best and second best results are \textbf{bold} and \underline{underlined}, respectively.}
        \vspace{-0.1in}
        \scalebox{0.98}{
        \centering
		\begin{tabular}{c|c|c|c|c||c|c|c|c}
			\hline
			Dataset&\multicolumn{4}{c||}{CoraFull}&\multicolumn{4}{c}{ogbn-arxiv}
			\\
			\hline
			 Way&\multicolumn{2}{c|}{2-way}&\multicolumn{2}{c||}{5-way}&\multicolumn{2}{c|}{2-way}&\multicolumn{2}{c}{5-way}\\\hline
 Shot & 1-shot&5-shot&1-shot&5-shot& 1-shot&5-shot& 1-shot&5-shot\\\hline\hline
 
MAML~\cite{finn2017model}&$50.90\pm2.30$&$56.19\pm2.37$&$22.63\pm1.19$&$27.21\pm1.32$&$58.16\pm2.35$&$65.10\pm2.56$&$27.36\pm1.48$&$29.09\pm1.62$
\\\hline
ProtoNet~\cite{snell2017prototypical}&$57.10\pm2.47$&$72.71\pm2.55$&$32.43\pm1.61$&$51.54\pm1.68$&$62.56\pm2.86$&$75.82\pm2.79$&$37.30\pm2.00$&$53
.31\pm1.71$\\\hline
Meta-GNN~\cite{zhou2019meta}&$75.28\pm3.85$&$84.59\pm2.89$&$55.33\pm2.43$&$70.50\pm2.02$&$62.52\pm3.41$&$70.15\pm2.68$&$27.14\pm1.94$&$31.52\pm1.7
1$\\\hline
GPN~\cite{ding2020graph}&$74.29\pm3.47$&$85.58\pm2.53$&$52.75\pm2.32$&$72.82\pm1.88$&$64.00\pm3.71$&$76.78\pm3.50$&$37.81\pm2.34$&$50.50\pm2.13$\\
\hline
AMM-GNN~\cite{wang21AMM}&$77.29\pm3.40$&$88.66\pm2.06$&$58.77\pm2.49$&$75.61\pm1.78$&$64.68\pm3.13$&$78.42\pm2.71$&$33.92\pm1.80$&$48.94\pm1.87$\\
\hline
G-Meta~\cite{huang2020graph}&$\underline{78.23\pm3.41}$&$\underline{89.49\pm2.04}$&$\underline{60.44\pm2.48}$&$\underline{75.84\pm1.70}$&$63.03\pm3.32$&$76.56\pm2.89$&$31.48\pm1.70$&$47.16\pm1.7
3$\\\hline
TENT~\cite{wang2022task}&$77.75\pm3.29$&$88.20\pm2.61$&$55.44\pm2.08$&$70.10\pm1.73$&$\underline{70.30\pm2.85}$&$\underline{81.35\pm2.77}$&$\underline{48.26\pm1.73}$&$\underline{61.38\pm1.72}$\\
\hline
COSMIC (\textbf{Ours.})&$\mathbf{84.32\pm2.75}$  & $\mathbf{94.51\pm2.47}$&$\mathbf{74.93\pm 2.49}$ &$\mathbf{86.34\pm2.17}$ &$\mathbf{75.71\pm3.17}$&$\mathbf{85.19\pm2.35}$&$\mathbf{53.28\pm2.19}$&$\mathbf{65.42\pm1.69}$\\
\hline
\end{tabular}}
\vspace{-0.05in}
		\label{tab:all_result}
	\end{table*}

	\begin{table*}[htbp]
			   					\setlength\tabcolsep{4.6pt}
		\centering
				\renewcommand{\arraystretch}{1.2}
        \scalebox{0.98}{
        \centering
		\begin{tabular}{c|c|c|c|c||c|c|c|c}
			\hline
			Dataset&\multicolumn{4}{c||}{Coauthor-CS}&\multicolumn{4}{c}{DBLP}
			\\
			\hline
 			 Way&\multicolumn{2}{c|}{2-way}&\multicolumn{2}{c||}{5-way}&\multicolumn{2}{c|}{2-way}&\multicolumn{2}{c}{5-way}\\\hline
 Shot & 1-shot&5-shot&1-shot&5-shot& 1-shot&5-shot& 1-shot&5-shot\\\hline\hline
 
MAML~\cite{finn2017model}&$56.90\pm2.41$&$66.78\pm2.35$&$27.98\pm1.42$&$42.12\pm1.40$&$53.04\pm2.58$&$58.67\pm2.44$&$23.94\pm1.15$&$35.09\pm1.61$\\\hline
ProtoNet~\cite{snell2017prototypical}&$59.92\pm2.70$&$71.69\pm2.51$&$32.13\pm1.52$&$49.25\pm1.50$&$60.97\pm2.56$&$72.81\pm2.73$&$31.31\pm1.58$&$52.26\pm1.88$\\\hline
Meta-GNN~\cite{zhou2019meta}&$85.90\pm2.96$&$90.11\pm2.17$&$52.86\pm2.14$&$68.59\pm1.49$&$82.60\pm3.23$&$86.15\pm3.29$&$\underline{67.24\pm2.72}$&$72.15\pm2.40$\\\hline
GPN~\cite{ding2020graph}&$84.31\pm2.73$&$90.36\pm1.90$&$60.66\pm2.07$&$\underline{81.79\pm1.18}$&$79.55\pm3.46$&$85.85\pm2.61$&$59.38\pm2.40$&$75.46\pm1.87$\\\hline
AMM-GNN~\cite{wang21AMM}&$84.38\pm2.85$&$89.74\pm1.20$&$62.04\pm2.26$&$81.78\pm1.24$&$79.77\pm3.37$&$\underline{91.74\pm2.01}$&$64.92\pm2.69$&$\underline{78.68\pm2.00}$\\\hline
G-Meta~\cite{huang2020graph}&$84.19\pm2.97$&$91.02\pm1.61$&$59.68\pm2.16$&$74.18\pm1.29$&$80.46\pm3.29$&$88.53\pm2.36$&$63.32\pm2.70$&$75.82\pm2.11$\\\hline
TENT~\cite{wang2022task}&$\underline{87.85\pm2.48}$&$\underline{91.75\pm1.60}$&$\underline{63.70\pm1.88}$&$76.90\pm1.19$&$\underline{84.40\pm2.73}$&$90.05\pm2.34$&$61.56\pm2.23$&$74.84\pm2.04$\\\hline
COSMIC (\textbf{Ours.})&$\mathbf{90.29\pm2.30}$&$\mathbf{94.32\pm1.93}$&$\mathbf{68.21\pm1.63}$&$\mathbf{85.47\pm1.11}$&$\mathbf{92.35\pm2.52}$& $\mathbf{94.82\pm1.69}$&$\mathbf{76.52\pm2.24}$&$\mathbf{85.31\pm1.92}$\\
\hline
\end{tabular}}
		\label{tab:all_result2}
	\end{table*}

\subsection{Experimental Settings}
To achieve a fair comparison of COSMIC with competitive baselines, we 
 compare the experimental results with those of the state-of-the-art few-shot node classification methods, including \textbf{Prototypical Networks (ProtoNet)}~\cite{snell2017prototypical}, \textbf{MAML}~\cite{finn2017model}, \textbf{Meta-GNN}~\cite{zhou2019meta}, \textbf{GPN}~\cite{ding2020graph}, \textbf{AMM-GNN}~\cite{wang21AMM}, \textbf{G-Meta}~\cite{huang2020graph}, and \textbf{TENT}~\cite{wang2022task}. More detailed information about those baselines is given in Appendix~\ref{app:baseline}.


\subsection{Overall Evaluation Results}
In this section, we compare the overall results of our framework with all baseline methods on few-shot node classification. The results are presented in Table~\ref{tab:all_result}. Specifically, to evaluate our framework under different few-shot settings, we conduct the experiments with different values of $N$ and $K$ under the $N$-way $K$-shot setting.
For the evaluation metrics, following the common practice~\cite{tan2022transductive}, we utilize the averaged classification accuracy and $95\%$ confidence interval over ten repetitions for a fair comparison. From the results, we can obtain the following observations:
(1) COSMIC consistently outperforms other baselines in all datasets with different values of $N$ and $K$. The results strongly validate the superiority of our contrastive meta-learning framework on few-shot node classification.
    (2) The performance of all methods significantly degrades when a larger value of $N$ is presented (i.e., more classes in each meta-task), since a larger class set will increase the variety of classes and hence lead to difficulties in classification. Nevertheless, our framework encounters a less significant performance drop compared with other baselines. This is because our contrastive meta-learning strategy is capable of handling various classes by leveraging both supervision and structural information in each episode.
    (3) With a larger value of $K$ (i.e., more support nodes
in each class), all methods exhibit decent performance improvements. Moreover, on Coauthor-CS, compared with the second-best baseline results (underlined), our framework achieves more significant improvements due to the enhancement of inter-class generalizability, which is crucial in Coauthor-CS with noticeably fewer meta-training classes than other datasets.
    (4) The confidence interval is generally larger on the setting with $K=1$, i.e., the 1-shot setting. This is because each meta-task only consists of one support node for each of the $N$ classes, making the decision boundary easy to overfit. In consequence, the results inevitably maintain a larger variance compared with the 5-shot setting. Nevertheless, our framework achieves the best results with comparable or smaller variance in comparison to baselines.
    
\vspace{-0.17cm}
    \subsection{Ablation Study}
\begin{figure}[!t]
   \vspace{-0.2cm}
		\centering
\captionsetup[sub]{skip=-1pt}
\subcaptionbox{ogbn-arxiv}
{\includegraphics[width=0.235\textwidth]{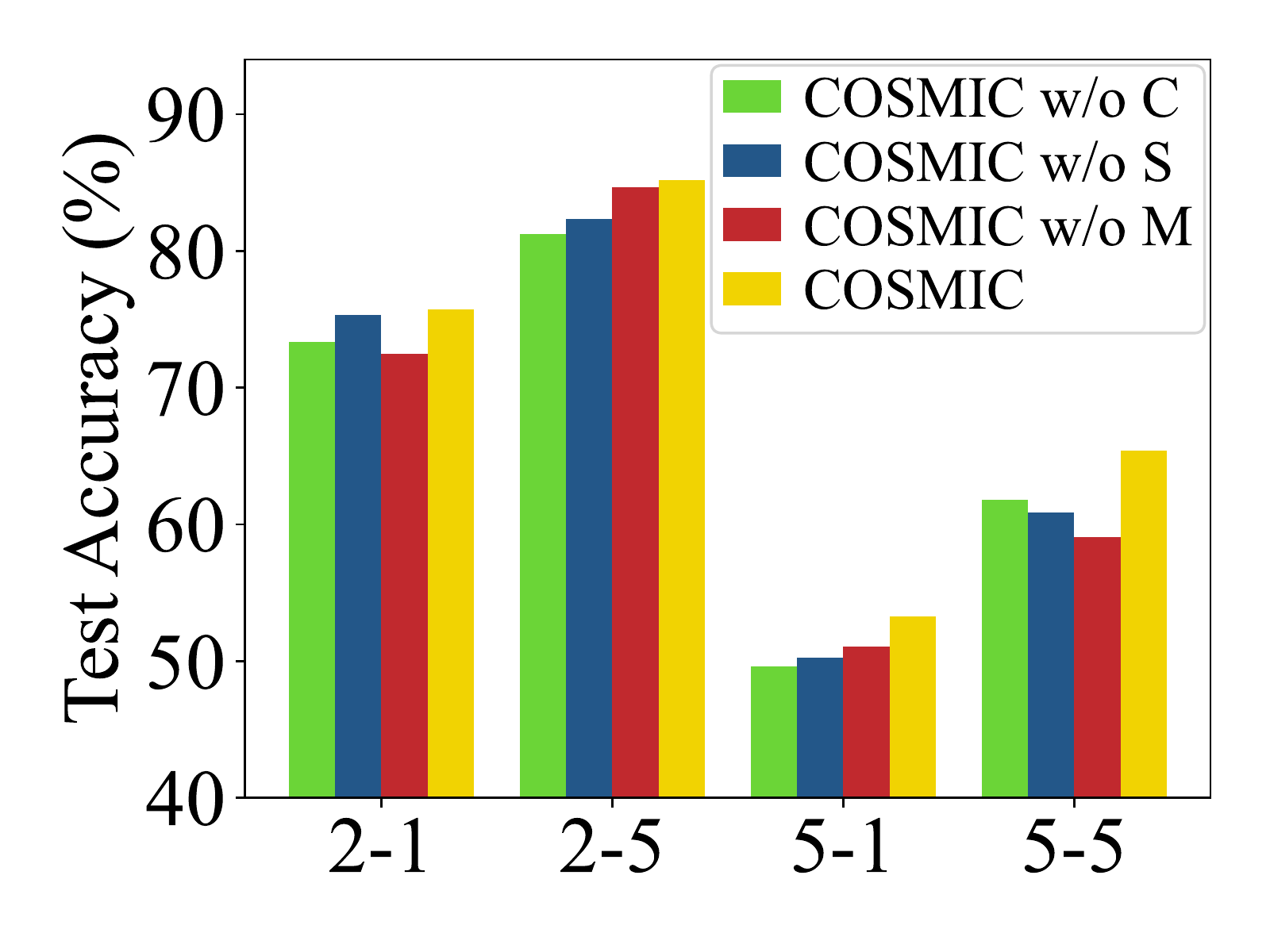}}
\subcaptionbox{DBLP}
{\includegraphics[width=0.235\textwidth]{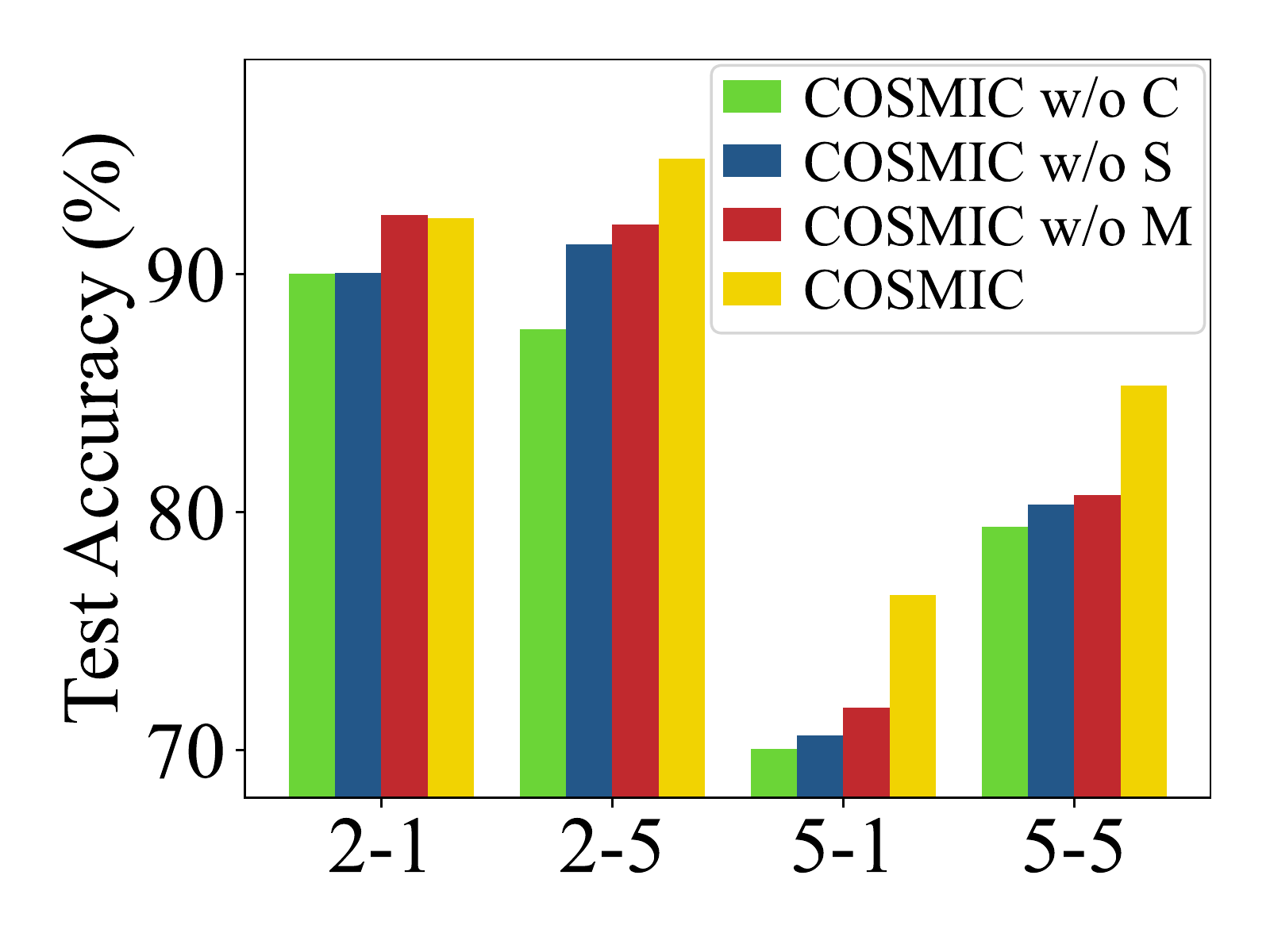}}
\vspace{-0.1in}
		\caption{Ablation study of our framework under different $N$-way $K$-shot settings (e.g., 2-1 denotes $2$-way $1$-shot) on ogbn-arxiv and DBLP. C refers to the contrastive meta-learning loss, S means representing nodes with subgraphs, and M denotes the similarity-sensitive mix-up.}
		\label{fig:ablation}
	\vspace{-0.3cm}
	\end{figure}

    In this section, we conduct an ablation study to evaluate the effectiveness of three modules in our framework COSMIC. In particular, we compare COSMIC with its three degenerate variants:
    (1) COSMIC without contrastive meta-learning (referred to as COSMIC w/o C). In this variant, we remove the contrastive learning loss so that only the cross-entropy loss is used for model training. 
    (2) COSMIC without using subgraphs (referred to as COSMIC w/o S). In this variant, we only leverage node representations learned from the same original graph.
    As a result, the model cannot incorporate contextual information of each node into meta-tasks.
    (3) COSMIC without similarity-sensitive mix-up (referred to as COSMIC w/o M). In this variant, the additional pairs of subgraphs are not incorporated. In consequence, the model cannot effectively achieve inter-class generalizability. 
    Fig.~\ref{fig:ablation} presents the results of our ablation study on ogbn-arxiv and DBLP datasets (similar results are observed on other datasets). Based on the results, we can achieve the following findings: (1) In general, our proposed framework COSMIC outperforms all three variants, which validates the effectiveness and necessity of the proposed three key components. Moreover, the advantage of the proposed framework becomes more significant on harder few-shot node classification tasks (i.e. a larger value of $N$ or a smaller value of $K$). This demonstrates the robustness of our framework regarding different $N$-way $K$-shot settings. 
        (2)~It can be observed that the variant without the contrastive meta-learning loss (i.e. COSMIC w/o C) generally exhibits inferior performance. This validates that our contrastive meta-learning strategy, which integrates a contrastive learning objective into each episode, succeeds in enhancing the generalizability of GNN models to meta-test classes.
        (3) The other two modules are also crucial for our framework. More specifically, when the value of $K$ decreases (i.e., fewer labeled nodes for each class in meta-tasks), the performance improvement brought by the subgraphs is more significant. This verifies that incorporating more contextual information in each meta-task can further compensate for the scarcity of labeled nodes. We further validate these points through experiments in the following section.

\vspace{-0.1cm}
 \subsection{Embedding Analysis}
 To explicitly illustrate the advantage of the proposed framework and the effectiveness of each designed component, in this subsection, we analyze the quality of the learned representations for nodes in meta-test classes through different training strategies. 
 
\subsubsection{Node Embedding Visualization via t-SNE} 
\begin{figure}[htbp]
		\centering
  \captionsetup[sub]{skip=-0.1pt}
	\subcaptionbox{COSMIC}{
\includegraphics[width=0.15\textwidth]{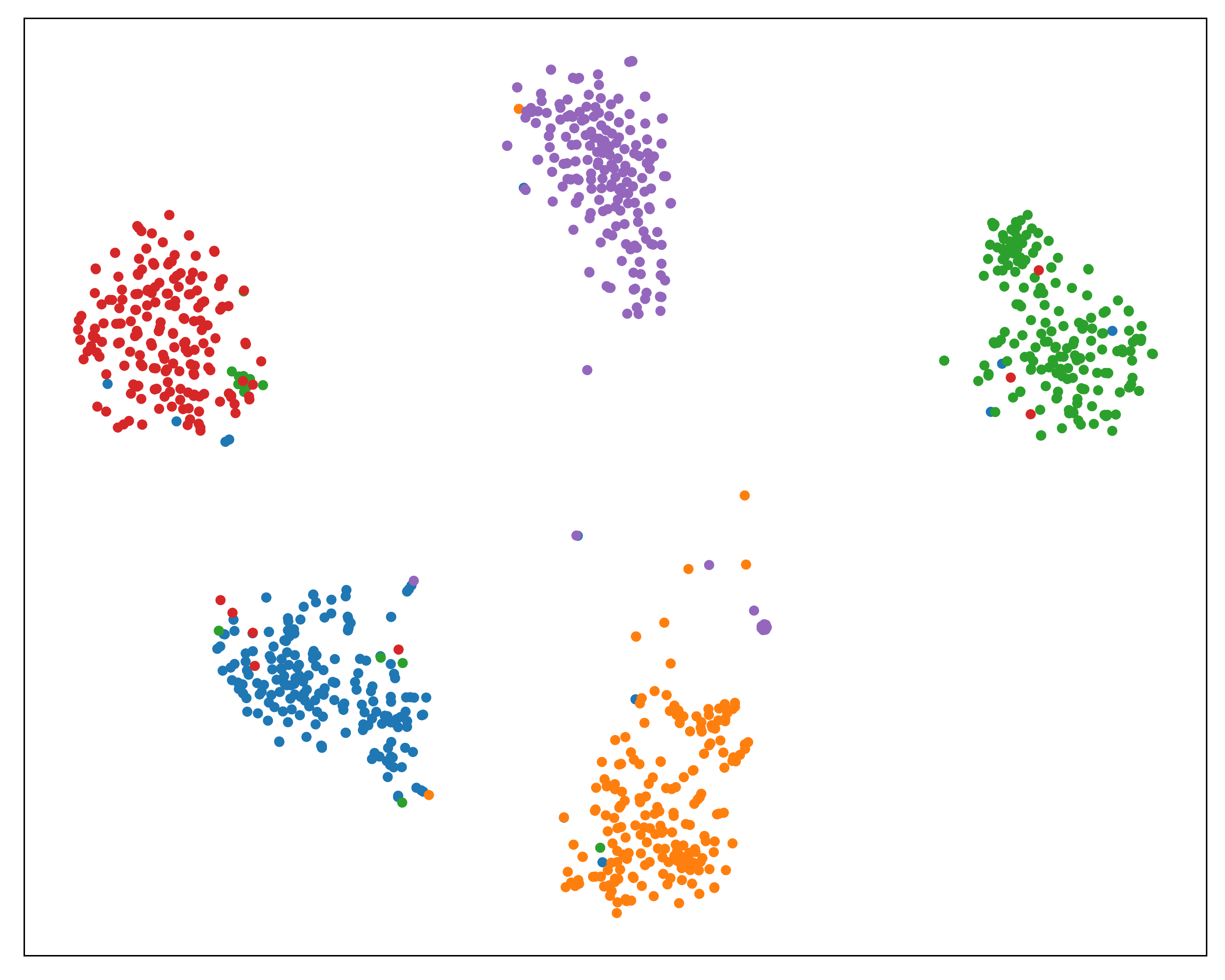}}
			\subcaptionbox{COSMIC w/o C}{
\includegraphics[width=0.15\textwidth]{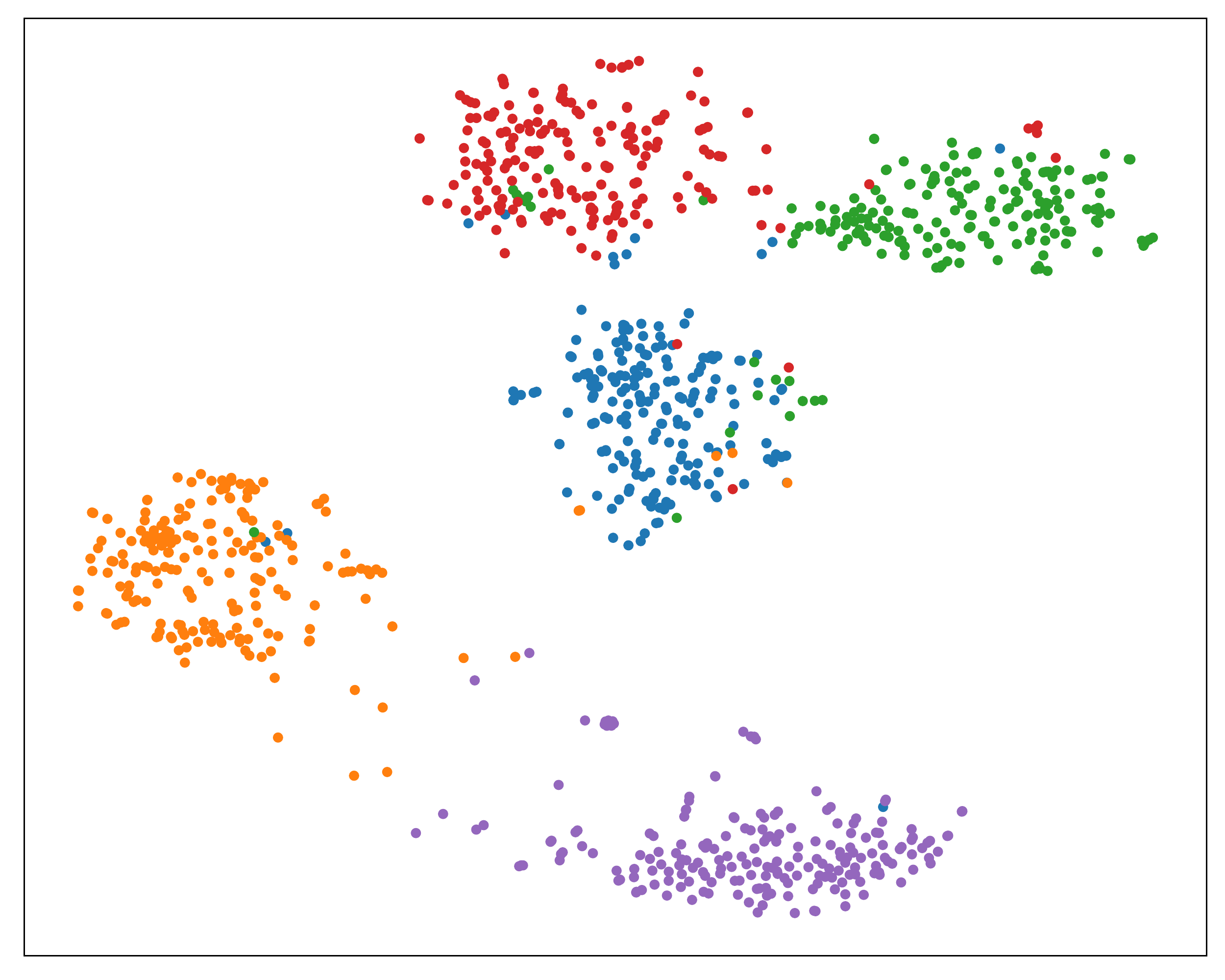}}
	\subcaptionbox{COSMIC w/o M}{
\includegraphics[width=0.15\textwidth]{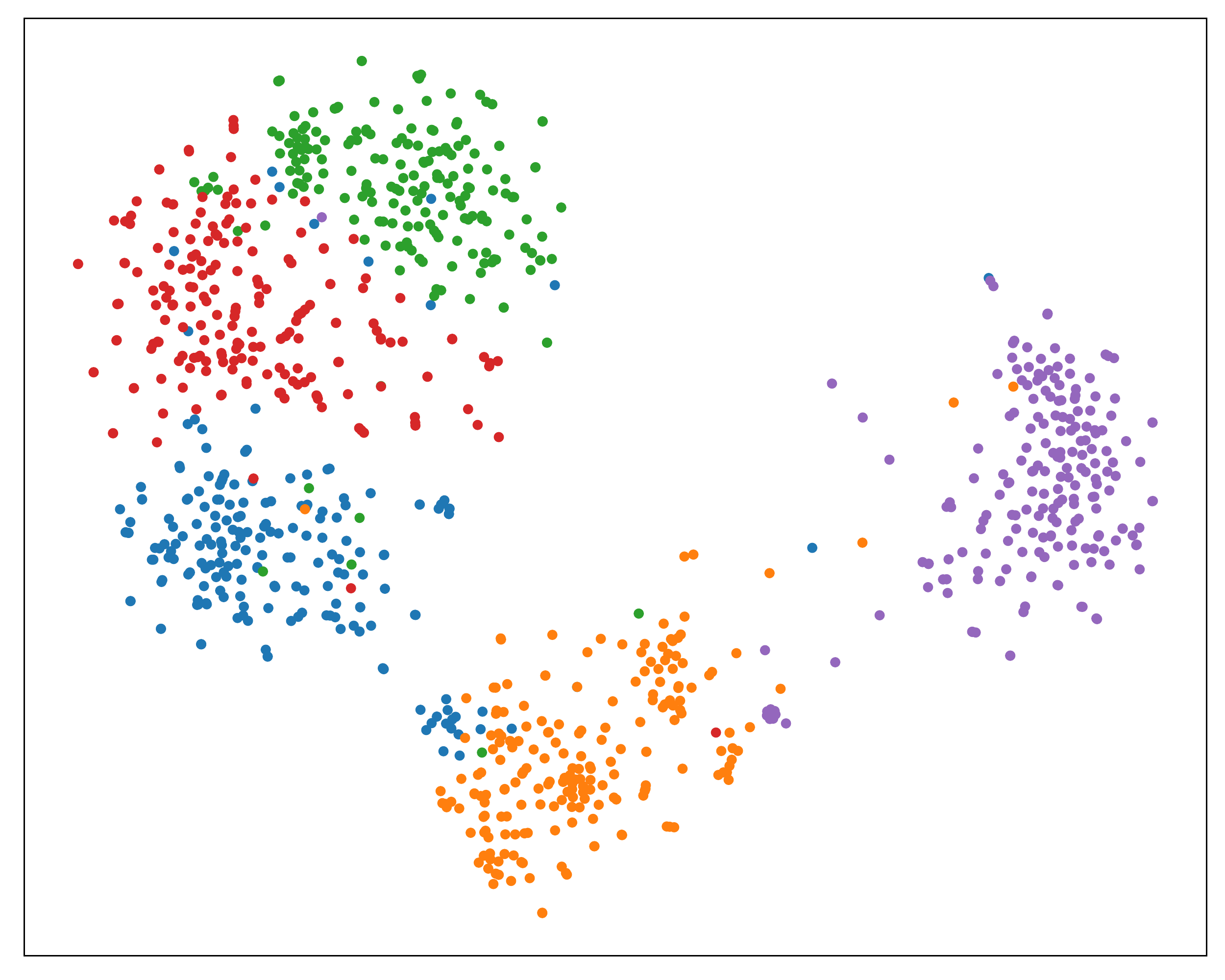}}

	\subcaptionbox{Meta-GNN}{
\includegraphics[width=0.15\textwidth]{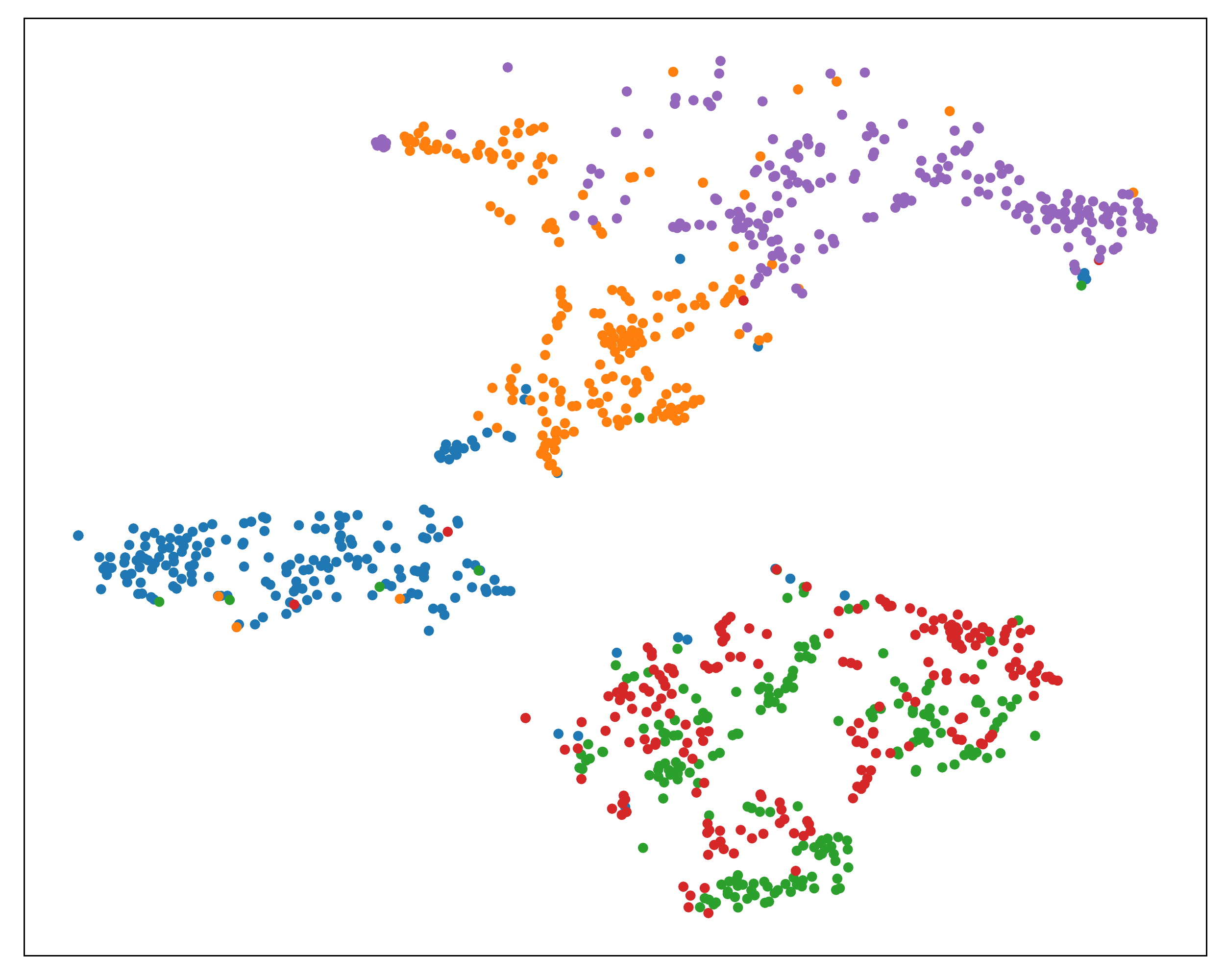}}
	\subcaptionbox{GPN}{
\includegraphics[width=0.15\textwidth]{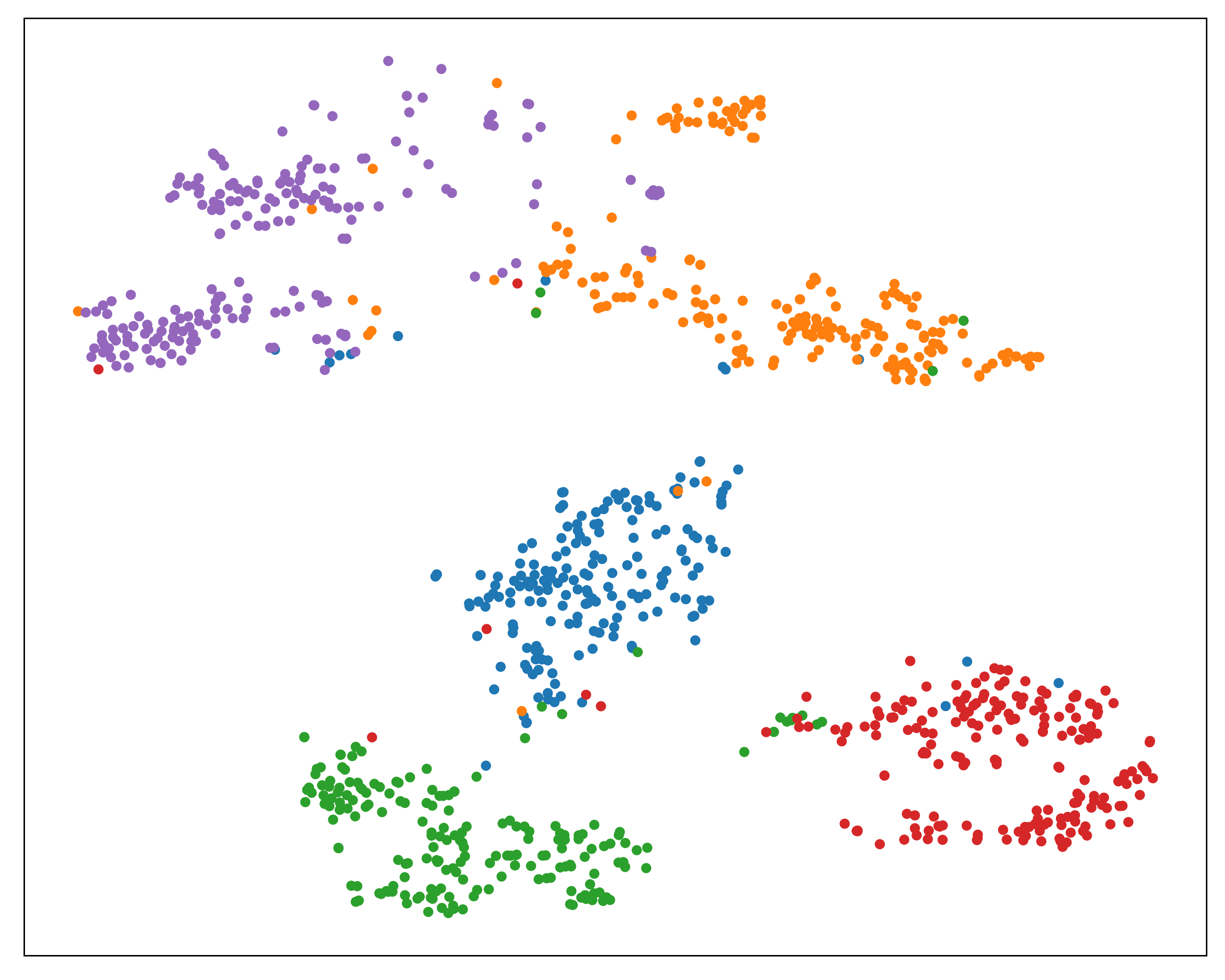}}
	\subcaptionbox{TENT}{
\includegraphics[width=0.15\textwidth]{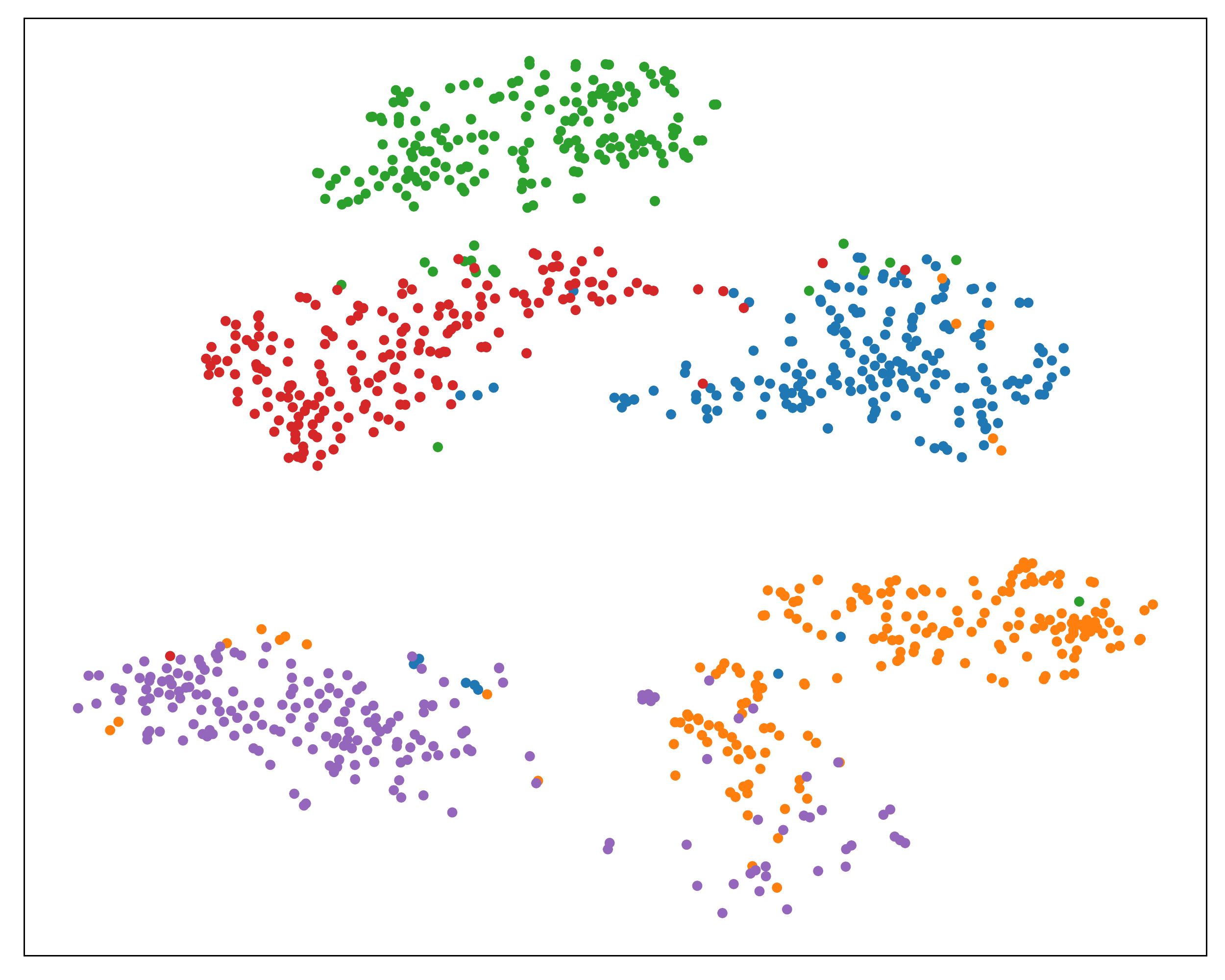}}
\vspace{-0.1in}
\caption{The visualization results on CoraFull under the 5-way 5-shot setting. Fig. (a)-(c) are from COSMIC under different ablation settings. Fig. (d)-(f) are obtained via baselines.}
\vspace{-0.3cm}
\label{fig:tsne}
	\end{figure}

We present the t-SNE visualization results of COSMIC on 5 randomly sampled meta-test classes from the CoraFull dataset in Fig.~\ref{fig:tsne}, in comparison to its ablated counterparts and other baselines. Based on the results, we can obtain the following discoveries: (1)
Comparing Fig.~\ref{fig:tsne} (a) to (d)-(f), we can observe that our proposed framework COSMIC can generate the most discriminative node embeddings on meta-test classes, compared to those competitive baselines: Meta-GNN, GPN, and TENT. This signifies that our framework extracts more generalizable knowledge and effectively transfers it to meta-test classes.
    (2) Comparing Fig.~\ref{fig:tsne} (a) to (b), it can be observed that node embeddings learned from COSMIC without the proposed contrastive meta-learning method will exhibit less intra-class discrimination. Comparing Fig.~\ref{fig:tsne} (a) to (c), we can observe that the learned node embeddings from different classes have more overlappings, which means the inter-class generalizability is limited. In other words, this visualization further validates the effectiveness of the proposed components for improving the GNN model's intra-class and inter-class generalizability.

\subsubsection{Node Embedding Clustering Evaluation}
For a more quantitive comparison, in this subsection, we present the detailed node embedding evaluations on CoraFull with NMI and ARI scores in Table~\ref{tab:nmi_result}. Similar to the previous experiments, we discover that the proposed framework, COSMIC, can learn the most discriminative node representations on meta-test classes, and the performance of ablated variants of COSMIC will degrade due to the limitations of both intra-class and inter-class generalizability.

	\begin{table}[htbp]
		\setlength\tabcolsep{8.5pt}
		\centering
		\renewcommand{\arraystretch}{1.2}
		\caption{The overall NMI ($\uparrow$) and ARI ($\uparrow$) results of COSMIC and baseline on two datasets under the 5-way 5-shot setting.}
        \vspace{-0.05in}
        \scalebox{0.95}{
		\begin{tabular}{c||c|c|c|c}
			\hline
			Dataset&\multicolumn{2}{c|}{\text{CoraFull}}&\multicolumn{2}{c}{\text{ogbn-arxiv}}
			\\
			\hline
						Metrics&\multicolumn{1}{c|}{NMI}&\multicolumn{1}{c|}{ARI}&\multicolumn{1}{c|}{NMI}&\multicolumn{1}{c}{ARI}\\
					\hline
			MAML~\cite{finn2017model}&$0.1622$&$0.0597$&$0.1732$&$0.1544$\\\hline
			ProtoNet~\cite{snell2017prototypical}&$0.2669$&$0.1263$&$0.2348$&$0.2132$\\\hline
   			Meta-GNN~\cite{zhou2019meta}&$0.5534$&$0.4196$&$0.2649$&$0.2214$\\\hline
      			GPN~\cite{ding2020graph}&$0.6001$&$0.4599$&$0.2505$&$0.2418$\\\hline
			AMM-GNN~\cite{wang21AMM}&$0.6247$&$0.5087$&$0.3029$&$0.2630$\\\hline
			G-Meta~\cite{huang2020graph}&$0.5003$&$0.3702$&$0.2430$&$0.2209$\\\hline
			TENT~\cite{wang2022task}&$0.5760$&$0.4652$&$0.3507$&$0.3113$\\\hline

      COSMIC w/o C&0.7505&0.6832&0.4436&0.2912    \\\hline
      COSMIC w/o M&0.7639&0.6958&0.4752&0.3277\\\hline
         			COSMIC&$\mathbf{0.7878}$&$\mathbf{0.7105}$&$\mathbf{0.4822}$&$\mathbf{0.3390}$\\\hline
\end{tabular}}
		\label{tab:nmi_result}
	\end{table}

    \subsection{Effect of Subgraph Size $K_s$}
\begin{figure}[t]
		\centering
\captionsetup[sub]{skip=-1pt}
\subcaptionbox{ogbn-arxiv}
{\includegraphics[width=0.235\textwidth]{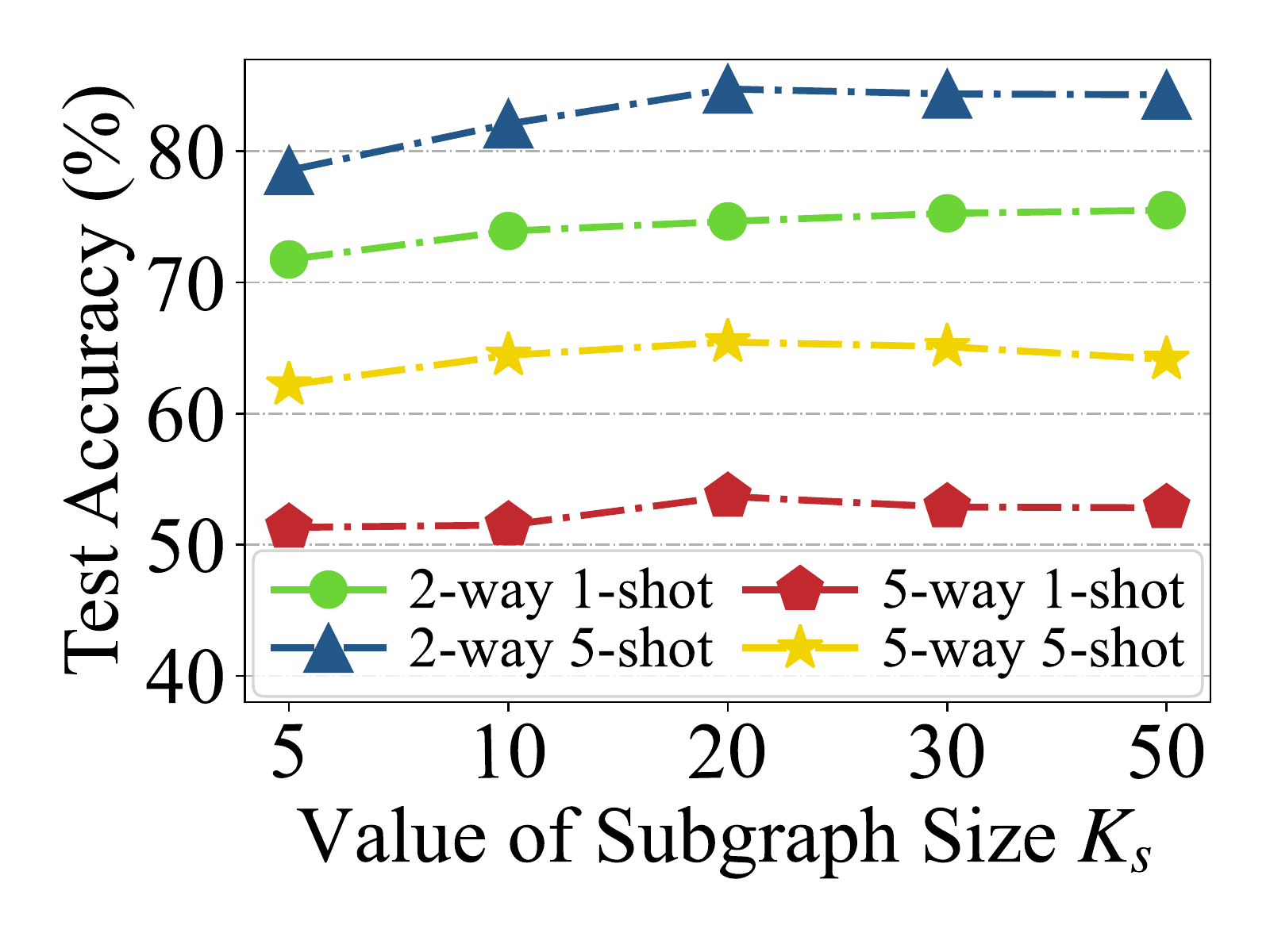}}
\subcaptionbox{DBLP}
{\includegraphics[width=0.235\textwidth]{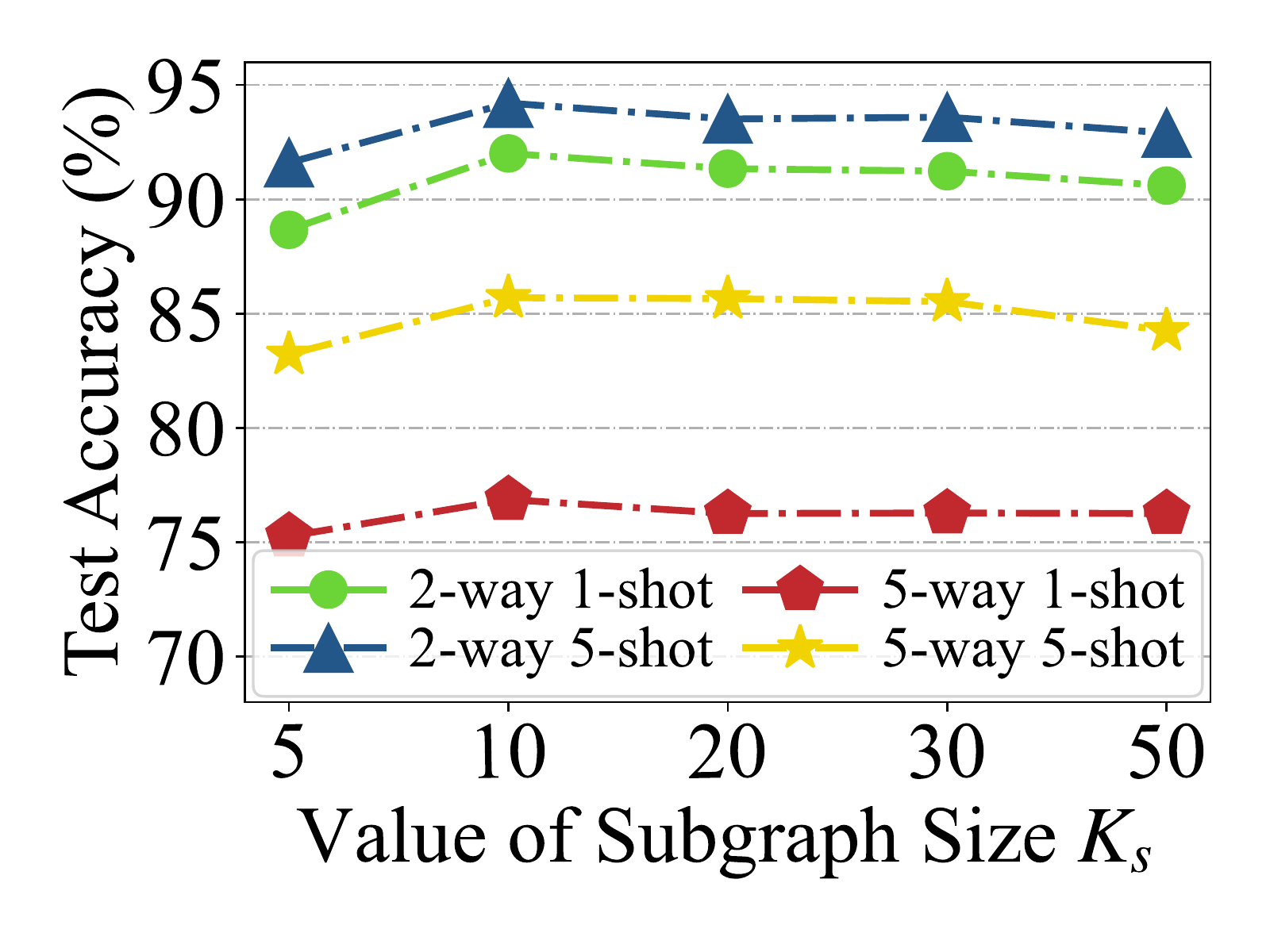}}
\vspace{-0.1in}
		\caption{The results of COSMIC with varying values of the subgraph size $K_s$ on ogbn-arxiv and DBLP.}
		\label{fig:subgraph_size}
\vspace{-0.12in}
	\end{figure}

    In this subsection, we conduct experiments to study the impact of the subgraph size $K_s$ in COSMIC. Specifically, with a larger value of $K_s$, our framework will incorporate more contextual information in each meta-task, which can further enhance the learning of transferable knowledge. 
    Fig.~\ref{fig:subgraph_size} reports the results of our framework with varying values of $K_s$ under four different few-shot settings. From the results, we can observe that increasing the size of subgraphs will first lead to better performance for our framework and then bring a slight performance drop. This is because larger subgraphs can involve more contextual information in each meta-task and thus contribute to the learning of transferable knowledge, while an excessively large subgraph can involve irrelevant information that harms the performance. Moreover, the performance advancement with larger subgraphs is more significant in 2-way settings, which means the incorporation of contextual information is more crucial in meta-tasks with less labeled nodes (i.e., a smaller support set).

\label{app:time}

    \subsection{Choice of Encoder}
    In this subsection, we conduct experiments on our framework with different choices of the encoder $\text{GNN}_\theta$. Notably, our framework does not require a specific implementation of GNNs and is thus compatible with any kind of GNN architectures. In particular, we change the GNN encoder to GAT~\cite{velivckovic2017graph}, GIN~\cite{xu2018powerful}, GraphSAGE~\cite{hamilton2017inductive} (denoted as SAGE in Table~\ref{tab:encoders}), and SGC~\cite{wu2019simplifying} to evaluate the effects of different GNNs. The results are provided in Table~\ref{tab:encoders}. 
    From the results, we can observe that generally GAT and SGC maintain relatively better performance on different few-shot settings. This is probably due to the fact that these GNN encoders can more effectively exploit the structural information, which can benefit from our contrastive meta-learning loss and mix-up strategy. Moreover, the results demonstrate that our proposed framework COSMIC maintains decent performance with various choices of GNN encoders, which validates the capability of COSMIC under different application scenarios. In other experiments, for the sake of simplicity and generality, we deploy GCN as the encoder for all the baselines and our proposed framework. 
    
        	\begin{table}[htbp]
		\setlength\tabcolsep{3.5pt}
		\centering
		\renewcommand{\arraystretch}{1.2}
		\caption{Comparisons of different GNNs used in our framework on ogbn-arxiv. }
        \vspace{-0.1in}
                \scalebox{0.95}{
		\begin{tabular}{c|c|c|c|c}
		\hline
        \multirow{2}*{Encoder}&\multicolumn{2}{c|}{2-way}&\multicolumn{2}{c}{5-way}\\\cline{2-5}
        &1-shot&5-shot&1-shot&5-shot\\\hline
        GCN~\cite{kipf2017semi}&$75.71\pm3.17$&$85.19\pm2.35$&$53.28\pm2.19$&$65.42\pm1.69$\\\hline
GAT~\cite{velivckovic2017graph}&$76.29\pm2.98$&$87.76\pm2.52$&$54.57\pm2.11$&$66.97\pm1.72$\\\hline
GIN~\cite{xu2018powerful}&$74.68\pm3.06$&$85.05\pm1.93$&$53.03\pm2.10$&$65.69\pm1.85$\\\hline
SAGE~\cite{hamilton2017inductive}&$73.92\pm3.13$&$83.29\pm2.21$&$51.72\pm2.11$&$63.74\pm1.77$\\\hline
SGC~\cite{wu2019simplifying}&$77.45\pm2.82$&$87.51\pm2.32$&$53.38\pm2.01$&$67.51\pm1.61$\\\hline

		\end{tabular}}
		\label{tab:encoders}
  \vspace{-0.2cm}
	\end{table}
 
\section{Related Work}

\subsection{Few-shot Node Classification}
Despite many breakthroughs in applying Graph Neural Networks (GNNs) to the node classification task~\cite{hamilton2017inductive,velivckovic2017graph,wang2022faith}, more recently, many studies~\cite{zhou2019meta,ding2020graph, wang2021reform} have shown that the performance of GNNs will severely degrade when the number of labeled node is limited, i.e., the few-shot node classification problem. Inspired by how humans transfer previously learned knowledge to new tasks, researchers propose to adopt the meta-learning paradigm~\cite{finn2017model} to deal with this label shortage issue~\cite{wang2022xfnc}. Particularly, the GNN models are trained by explicitly emulating the test environment for few-shot learning, where the GNNs are expected to gain the adaptability to generalize onto new domains. For example, Meta-GNN \cite{zhou2019meta} applies MAML \cite{finn2017model} to learn directions for optimization with limited labels. GPN~\cite{ding2020graph} adopts Prototypical Networks \cite{snell2017prototypical} to make the classification based on the distance between the node feature and the prototypes. MetaTNE~\cite{lan2020node} and RALE \cite{liu2021relative} also use episodic meta-learning to enhance the adaptability of the learned GNN encoder and achieve similar results. However, those existing works usually directly apply meta-learning to graphs~\cite{wang2022glitter}, ignoring the crucial distinction from images that nodes in a graph are not i.i.d. data, thus leading to several drawbacks as discussed in the paper. Our work bridges the gap by developing a novel contrastive meta-learning framework for few-shot node classification.

\subsection{Graph Contrastive Learning} Contrastive learning has become an effective representation learning paradigm in image~\cite{chen2020simple}, text~\cite{wang2021cline}, and graph~\cite{hassani2020contrastive,zhu2021graph,ding2022data} domains. Specifically, for a typical graph contrastive learning method, a GNN encoder is forced to maximize the consistency between differently augmented views for original graph data. The augmentation is achieved by specific heuristic transformations, such as randomly dropping edges and nodes~\cite{hassani2020contrastive,you2020graph}, and randomly perturbing attributes of nodes and edges~\cite{tong2021directed}. Pretraining with such general pretexts will help the GNN model to learn transferable graph patterns~\cite{hassani2020contrastive}. The pretrained model parameters have been proven to be a superior initialization for GNNs when fine-tuned on various downstream tasks, including node classification~\cite{hassani2020contrastive,you2020graph}. However, all existing works fine-tune the GNNs on sufficiently labeled datasets, making them unsuitable for scenarios where there are only a few labeled nodes for fine-tuning~\cite{tan2022graph}. 

\subsection{Mix-up on Graphs}
Mix-up~\cite{zhang2018mixup} has become a popular data augmentation technique for training deep models to enhance their generalizability and robustness. As a common practice~\cite{zhang2018mixup,verma2019manifold,guo2019mixup}, both the attributes and labels of a pair of original instances are linearly interpolated and integrated into the original datasets to train the model, with weights sampled from Beta distributions. To achieve this manner for graphs, one work~\cite{wang2021mixup} modifies graph convolution to mix the graph parts within the receptive field. Another work~\cite{han2022g} proposes to learn a graph generator to align the pair of graphs and interpolate the generated counterparts. However, these methods require extra deep modules to learn, making the generated graphs hard to interpret. In our work, we design a novel heuristic method that calculates subgraph-level similarities, based on which the mix-up strategy can adaptively generate subgraphs with additional classes. We incorporate them into the novel contrastive meta-learning framework. 

\section{Conclusion}
In this paper, we investigate the few-shot node classification problem, which aims to assign labels for nodes with only limited labeled nodes as references. We improve the meta-learning strategy from the perspective of enhancing the intra-class and inter-class generalizability for the learned node embeddings. Towards this end, we propose a contrastive meta-learning framework, COSMIC, with the following novel designs: (1) we propose a two-step optimization based on graph contrastive learning, which utilizes subgraphs to represent nodes to learn more comprehensive context structural information and enhances intra-class generalizability. (2) We put forward a similarity-sensitive mix-up method that generates additional hard negative classes for the graph contrastive step to strengthen the inter-class generalizability of the learned node embeddings. Comprehensive experiments on four real-world datasets validate the effectiveness of our proposed framework. However, there still exist several unresolved problems. For instance, our framework still requires sufficiently labeled nodes in meta-training classes. Future work can extend our framework to scenarios where no or limited labeled nodes exist.

  \section{Acknowledgements}
This work is supported by the National Science Foundation under grants (IIS-2006844, IIS-2144209, IIS-2223769, IIS-2229461, CNS-2154962, and BCS-2228534), the Commonwealth Cyber Initiative awards (VV-1Q23- 007 and HV-2Q23-003), the JP Morgan Chase Faculty Research Award, the Cisco Faculty Research Award, the Jefferson Lab subcontract 23-D0163, and the UVA 4-VA collaborative research grant.
	
		\bibliographystyle{ACM-Reference-Format}
		\bibliography{acmart}
			\clearpage
\appendix
\section{Notations}\label{app:Notations}
In this section, we provide used notations in this paper along with their descriptions for comprehensive understanding.
\begin{table}[htbp]
\small
\setlength\tabcolsep{5pt}
\caption{Notations used in this paper.} 

\label{tb:symbols}
\begin{tabular}{cc}

\hline

\textbf{Notations}       & \textbf{Definitions or Descriptions} \\
\hline

$G$   &  the input graph\\
$\mathcal{V}$, $\mathcal{E}$  & the node set and the edge set of $G$\\
$\X$ & the input node features of $G$\\
$\mathcal{C}_{tr}$,$\mathcal{C}_{te}$ & the meta-training set and the meta-test class set\\
$\mathcal{T}$, $\mathcal{S}$, $\mathcal{Q}$&a meta-task and its support set and query set\\
$N$&the number of support classes in each meta-task\\
$K$&the number of labeled nodes in each class\\
$K_s$&the size of the sampled subgraph for each node\\
$\eta_{MC}$&the learning rate for contrastive meta-learning loss\\
$\eta_{CE}$&the learning rate for cross-entropy loss\\
$\Lambda_A$& mixing ratio matrices for adjacency matrices\\
$\Lambda_X$&mixing ratio matrices for node features\\
\hline
\end{tabular}
\end{table}
\section{Reproducibility}\label{app:Reproducibility}

\subsection{Baseline}\label{app:baseline}
We conduct experiments with the following baseline methods to
compare performance:
\begin{itemize}
    \item \textbf{Prototypical Networks (ProtoNet)}~\cite{snell2017prototypical}: ProtoNet learns prototypes for classes within each meta-task and classifies query instances via their similarities to prototypes.
        \item \textbf{MAML}~\cite{finn2017model}: MAML proposes to optimize model parameters according to gradients on the support instances and meta-update parameters based on query instances.
    \item \textbf{Meta-GNN}~\cite{zhou2019meta}: Meta-GNN combines MAML and Graph Neural Networks (GNNs) to perform meta-learning on graph data for few-shot node classification.
    \item \textbf{GPN}~\cite{ding2020graph}: GPN learns node importance and combines Prototypical Networks to improve performance.
    \item \textbf{AMM-GNN}~\cite{wang21AMM}: AMM-GNN proposes to extend MAML with an attribute matching mechanism.
        \item \textbf{G-Meta}~\cite{huang2020graph}: G-Meta utilizes subgraphs to learn node representations, based on which the classification is conducted.
    \item \textbf{TENT}~\cite{wang2022task}: TENT proposes to reduce the task variance among various meta-tasks and conduct task-adaptive few-shot node classification from different levels.
    \end{itemize}

\subsection{Experimented Datasets}\label{app:dataset}
To evaluate the performance of COSMIC on few-shot node classification, we conduct experiments on four prevalent real-world benchmark node classification datasets: CoraFull~\cite{bojchevski2018deep}, ogbn-arxiv~\cite{hu2020open}, Coauthor-CS~\cite{shchur2018pitfalls}, and DBLP~\cite{tang2008arnetminer}. Their statistics are provided in Table~\ref{tab:statistics}.
\begin{table}[htbp]
	
		\setlength\tabcolsep{6.2pt}
		\small
		\centering
		\renewcommand{\arraystretch}{1.2}
		\caption{Statistics of four node classification datasets. }
		\begin{tabular}{ccccc}
		\hline
        Dataset&\# Nodes & \# Edges & \# Features & Class Split\\
        \hline
        CoraFull&19,793&63,421&8,710&40/15/15\\
        ogbn-arxiv&169,343&1,166,243&128&20/10/10\\
        Coauthor-CS&18,333&81,894&6,805&5/5/5\\        DBLP&40,672&144,135&7,202&77/30/30\\
  \hline
		\end{tabular}
		\label{tab:statistics}
	\end{table}
\begin{itemize}

    \item \textbf{CoraFull}~\cite{bojchevski2018deep} is an extension of the prevalent dataset Cora~\cite{yang2016revisiting} from the entire citation network. On this graph, papers and citation relations are represented as nodes and edges, respectively. The classes of nodes are obtained according to the paper topic. For this
dataset, we use 40/15/15 node classes for meta-training/meta-validation/meta-test.
    \item \textbf{ogbn-arxiv}~\cite{hu2020open} is a citation network based on CS papers extracted from MAG~\cite{wang2020microsoft}. Specifically, it is a directed graph, where nodes represent CS arXiv papers while edges denote the citation relations between papers. The node labels are assigned according to the 40 CS subject areas in arXiv. For this dataset, we use 20/10/10 node classes for meta-training/meta-validation/meta-test.

    \item \textbf{Coauthor-CS}~\cite{shchur2018pitfalls} is a co-authorship graph obtained from the Microsoft Academic Graph in the KDD Cup 2016 challenge. Specifically, nodes represent the authors, and edges denote the relations that they co-authored a paper. Moreover, node features represent the paper keywords in each author’s papers. The node classes are assigned based on the most active fields of the authors. We use 5/5/5 node classes for meta-training/meta-validation/meta-test.

        \item \textbf{DBLP}~\cite{tang2008arnetminer} is a citation network, where the nodes represent papers, and edges denote the citation relations between papers. Specifically, the node features are obtained based on the paper abstract, and node classes are assigned according to the paper venues. For this dataset, we use 77/30/30 node classes for meta-training/meta-validation/meta-test.
    \end{itemize} 
\subsection{Implementation Details}
 In this section, we provide more details on the implementation settings of our experiments. 
	\label{appendix:implementation}
	Specifically, we implement COSMIC with PyTorch~\cite{paszke2017automatic} and train our framework on a single 48GB Nvidia A6000 GPU. We utilize a one-layer GCN~\cite{kipf2017semi} as our base GNN model with the hidden size set as 1024. In the beta distribution, we set the constant value $C$ as 10 and $\beta$ as 5. Moreover, we set the number of meta-training tasks $T$ as 1000. During the meta-test phase, we randomly sample 100 meta-test tasks, where the query set size (i.e., $|\mathcal{Q}|$) is set as 10. We adopt the Adam~\cite{kingma2014adam} optimization method, where the learning rates for contrastive meta-learning loss and cross-entropy loss, $\eta_{MC}$ and $\eta_{CE}$, are both set as 0.001. 

\begin{table}[htbp]

		\setlength\tabcolsep{5pt}
		\centering
		\renewcommand{\arraystretch}{1.2}
		\caption{The overall training time results of COSMIC and baselines on two datasets under the 5-way 1-shot setting.}

        \scalebox{0.95}{
		\begin{tabular}{c||c|c|c|c}
			\hline
			Dataset&\multicolumn{2}{c|}{\text{Coauthor-CS}}&\multicolumn{2}{c}{\text{ogbn-arxiv}}
			\\
			\hline
						Result&\multicolumn{1}{c|}{Time (s)}&\multicolumn{1}{c|}{\# Episodes}&\multicolumn{1}{c|}{Time (s)}&\multicolumn{1}{c}{\# Episodes}\\
					\hline
			MAML~\cite{finn2017model}&19.7&298.0&22.1&145.7\\\hline
			ProtoNet~\cite{snell2017prototypical}&8.33&125.6&21.4&153.5\\\hline
   			Meta-GNN~\cite{zhou2019meta}&86.4&461.9&157.5&261.8\\\hline
      			GPN~\cite{ding2020graph}&9.4&112.1&35.2&257.2\\\hline
			AMM-GNN~\cite{wang21AMM}&211.09&264.5&237.0&193.3\\\hline
			G-Meta~\cite{huang2020graph}&285.5&362.8&314.8&117.1\\\hline
			TENT~\cite{wang2022task}&148.2&976.4&49.3&366.7\\\hline
         			COSMIC&105.5&231.8&33.9&141.5\\\hline
\end{tabular}}
		\label{tab:training_time}
	\end{table}

\vspace{0.1in}
\section{Training Time}
In this subsection, we conduct additional experiments to evaluate the computational cost of our framework compared to other baselines. Specifically, in Table~\ref{tab:training_time}, we demonstrate the overall training time of the proposed framework, COSMIC, and baselines on two typical datasets: Coauthor-CS, which is a simpler dataset with a smaller graph, and ogbn-arxiv, which is a harder dataset with a significantly larger graph. For all methods, we consider the training time till convergence (time for preprocessing excluded) for 10 runs and report their average. For consistency, we run all the experiments on a single 48 GB Nvidia A6000 GPU. As shown in the results, for the smaller and simpler graph dataset Coauthor-CS, the proposed COSMIC requires more training time to achieve higher accuracy. However, for the larger and harder graph dataset ogbn-arxiv, COSMIC can achieve the best performance while demanding 
similar or even less training time. This is because COSMIC learns the encoder at the subgraph level, while all the existing meta-learning-based baselines resort to training a global graph encoder, thus leading to more training episodes. This experiment demonstrates that the proposed COSMIC can scale well to large and complex graphs.

\begin{table}[!t]
 
		\setlength\tabcolsep{4pt}
		\centering
		\renewcommand{\arraystretch}{1.2}
		\caption{The overall results of using other meta-learning frameworks with our contrastive loss on ogbn-arxiv.}

        \scalebox{0.95}{
		\begin{tabular}{c||c|c|c|c}
			\hline
			 Way&\multicolumn{2}{c|}{2-way}&\multicolumn{2}{c}{5-way}\\\hline
 Shot & 1-shot&5-shot&1-shot&5-shot\\\hline
\hline
ProtoNet~\cite{snell2017prototypical} & 67.36$\pm$3.73 & 76.99$\pm$2.87 & 39.10$\pm$1.64 & 56.57$\pm$2.14 \\
Matching~\cite{vinyals2016matching} & 68.86$\pm$3.30 & 77.01$\pm$2.79 & 44.79$\pm$1.85 & 58.72$\pm$1.94 \\
Relation~\cite{sung2018learning} & 71.10$\pm$3.00 & 79.56$\pm$3.12 & 47.88$\pm$1.66 & 60.96$\pm$1.54 \\
COSMIC & 75.71$\pm$3.17 & 85.19$\pm$2.35 & 53.28$\pm$2.19 & 65.42$\pm$1.69 \\\hline
\end{tabular}}
		\label{tab:meta}
\vspace{-0.096in}
	\end{table}

\vfill\eject 
\section{Additionla Results}

\subsection{Meta-learning Frameworks}

In this subsection, we provide the results of other meta-learning frameworks with our contrastive learning loss,
as shown in Table~\ref{tab:meta}. From the results, we observe that using MAML is generally better than other meta-learning frameworks. Regarding the meta-learning framework, we chose to utilize MAML due to its ability to perform both inner- and outer-loop optimizations, which aligns with our strategy that first learns node representations with intra-class and inter-class generalizability and then conducts classification. Although our contrastive loss can also be incorporated into other meta-learning frameworks, they do not contain such a two-stage optimization, which makes them less effective for our purpose.

\subsection{Contrastive Leraning Frameworks}
 In this subsection, we conduct additional experiments and provide the results of different contrastive learning methods with MAML as the framework, as shown in Table~\ref{tab:contrast}.

From the results, we observe that our proposed contrastive meta-learning loss achieves superior performance. This is because compared to other contrastive learning methods, our strategy can enhance the intra-class and inter-class generalizability of the model, which is more suitable for few-shot node classification problems.

\begin{table}[!t]
		\setlength\tabcolsep{4pt}
		\centering
		\renewcommand{\arraystretch}{1.2}
		\caption{The overall results of using other contrastive learning frameworks with MAML on ogbn-arxiv.}
        \scalebox{0.95}{
		\begin{tabular}{c||c|c|c|c}
			\hline
			 Way&\multicolumn{2}{c|}{2-way}&\multicolumn{2}{c}{5-way}\\\hline
 Shot & 1-shot&5-shot&1-shot&5-shot\\\hline
\hline
MVGRL~\cite{hassani2020contrastive} & 62.20$\pm$3.14 & 74.45$\pm$3.27 & 40.63$\pm$1.66 & 56.54$\pm$1.84 \\
GraphCL~\cite{you2020graph}& 67.50$\pm$3.18 & 79.68$\pm$2.23 & 44.34$\pm$1.58 & 60.49$\pm$2.02 \\
GRACE~\cite{zhu2020deep} & 65.81$\pm$3.06 & 76.03$\pm$2.74 & 42.90$\pm$1.68 & 57.96$\pm$1.61 \\
MERIT~\cite{jin2021multi}& 66.58$\pm$2.84 & 79.18$\pm$3.15 & 47.56$\pm$1.36 & 59.19$\pm$1.70 \\
COSMIC (Ours.) & 75.71$\pm$3.17 & 85.19$\pm$2.35 & 53.28$\pm$2.19 & 65.42$\pm$1.69 \\
\hline
\end{tabular}}
		\label{tab:contrast}
    \vspace{-0.1in}
	\end{table}

\end{document}